\documentclass[reprint,amsmath,amssymb,aps,prx]{revtex4-2}
\usepackage{graphicx}
\usepackage{color}
\usepackage{amsmath}
\usepackage{amssymb}
\usepackage{amsthm}
\usepackage{extarrows}
\usepackage{booktabs}
\usepackage{float}
\usepackage[dvipdfm,colorlinks,urlcolor=blue,linkcolor=blue,anchorcolor=blue,citecolor=blue]{hyperref}

\usepackage[left]{lineno}

\usepackage{dcolumn}
\usepackage{bm}
\usepackage{multirow}
\begin{document}

\title{Parallel BiLSTM-Transformer networks for forecasting chaotic dynamics}

\author{Junwen Ma$^{1,2}$}
\author{Mingyu Ge$^{1}$}
\author{Yisen Wang$^{3,4}$}
\author{Yong Zhang$^{5}$}
\author{Weicheng Fu$^{1,2,3}$}
\email{fuweicheng@tsnu.edu.cn}

\affiliation{
$^1$ Department of Physics, Tianshui Normal University, Tianshui 741001, Gansu, China\\
$^2$ Key Laboratory of Atomic and Molecular Physics $\&$ Functional Material of Gansu Province, College of Physics and Electronic Engineering, Northwest Normal University, Lanzhou 730070, China\\
$^3$ Lanzhou Center for Theoretical Physics, Lanzhou University, Lanzhou 730000, Gansu, China\\
$^4$ Key Laboratory for Quantum Theory and Applications of the Ministry of Education, Lanzhou University, Lanzhou, Gansu 730000, China\\
$^5$ Department of Physics, Xiamen University, Xiamen 361005, Fujian, China
}

\date{\today}


\begin{abstract}
The nonlinear nature of chaotic systems results in extreme sensitivity to initial conditions and highly intricate dynamical behaviors, posing fundamental challenges for accurately predicting their evolution. To overcome the limitation that conventional approaches fail to capture both local features and global dependencies in chaotic time series simultaneously, this study proposes a parallel predictive framework integrating Transformer and Bidirectional Long Short-Term Memory (BiLSTM) networks. The hybrid model employs a dual-branch architecture, where the Transformer branch mainly captures long-range dependencies while the BiLSTM branch focuses on extracting local temporal features. The complementary representations from the two branches are fused in a dedicated feature-fusion layer to enhance predictive accuracy. As illustrating examples, the model's performance is systematically evaluated on two representative tasks in the Lorenz system. The first is autonomous evolution prediction, in which the model recursively extrapolates system trajectories from the time-delay embeddings of the state vector to evaluate long-term tracking accuracy and stability. The second is inference of unmeasured variable, where the model reconstructs the unobserved states from the time-delay embeddings of partial observations to assess its state-completion capability. The results consistently indicate that the proposed hybrid framework outperforms both single-branch architectures across tasks, demonstrating its robustness and effectiveness in chaotic system prediction.
\end{abstract}

\maketitle

\section{Introduction}\label{sec:level1}

Chaotic phenomena pervade diverse domains such as physics, meteorology, biology, and economics. Their inherent nonlinearity gives rise to extreme sensitivity to initial conditions and intricate dynamical behaviours, making accurate prediction of their evolution a long-standing challenge in scientific research~\cite{cvitanovic_universality_2017,ratas2024application}. Classical approaches to chaotic time series prediction rely primarily on numerical integration and reconstruction techniques rooted in nonlinear dynamics, including delay embedding~\cite{takens1981dynamical} and phase-space reconstruction~\cite{packard1980geometry}. However, these methods often fail to achieve reliable accuracy in high-dimensional systems or under irregular sampling conditions~\cite{hart2023estimating}.

In recent years, machine-learning methods, owing to their data-driven and model-free advantages, have gained increasing prominence in chaotic time series prediction. Among them, reservoir computing (RC) has proved particularly influential, offering efficient training, strong nonlinear modelling capacity, and successful applications across a wide range of chaotic systems~\cite{wei2022reservoir,moon2019temporal,zhong2022memristor,zhong2021dynamic,chen2023all,brunner2013parallel,kan2022physical,du2017reservoir,rafayelyan2020large,jaeger2004harnessing,milano2022materia,pathak2018model,yan2022reservoir,cucchi2021reservoir,pathak2017using,panahi2024adaptable,fang2025dynamical}.
Nevertheless, conventional RC frameworks still struggle to simultaneously capture long-range dependencies and fine-grained local structures, enhance nonlinear expressiveness, and maintain long-term dynamical consistency~\cite{xiao2021predicting}. To address these challenges, numerous RC variants have been developed, improving adaptability and generalisation in complex system prediction~\cite{gauthier2021next,2021-chaos,chepuri2024hybridizing,wang2024enhanced}.

The rapid development of deep learning has further expanded the toolkit for chaotic sequence modelling~\cite{ramadevi2022chaotic}. Recurrent neural networks (RNNs), particularly long short-term memory (LSTM) networks, have become mainstream due to their effective gating mechanisms that mitigate gradient pathologies and capture temporal dependencies within a moderate range~\cite{hochreiter1997long,landi2021working,shahi2022prediction}. The bidirectional LSTM (BiLSTM) further enhances contextual awareness by incorporating both forward and backward temporal information, achieving superior performance in time-sensitive tasks. However, both LSTM and BiLSTM remain limited in modelling ultra-long dependencies and global dynamical structures~\cite{Wang2023_FuzzyLSTM,bai2018empirical,tiezzi2025back}.

To overcome these limitations, attention-based architectures have been introduced~\cite{ashish2017attention}. The Transformer, leveraging self-attention mechanisms, eliminates recurrence and convolution, efficiently captures global dependencies, and employs positional encoding to represent temporal order---thus overcoming the long-distance modelling constraints of RNNs~\cite{ashish2017attention}. Variants such as Informer~\cite{zhou2021informer} further extend efficiency and expressive power to ultra-long sequences. However, while Transformer excels at global feature extraction, it tends to perform less effectively in capturing local dynamical variations and abrupt changes~\cite{wu2021autoformer,zeng2023transformers,wu2022timesnet}. In contrast, LSTM-based architectures remain well suited to modelling short-range temporal dynamics.
Motivated by this complementarity, recent studies have explored serial combinations of LSTM and Transformer architectures for real-world time series prediction~\cite{lim2021temporal,cao2024remaining,sci7010007,2025Guo}, including applications such as battery health forecasting~\cite{2025Guo}.

In this work, we propose a parallel dual-branch architecture for time series prediction, which integrates BiLSTM and Transformer modules. The Transformer branch mainly captures long-range dependencies through self-attention, whereas the BiLSTM branch focuses on extracting local dynamical patterns. Their outputs are fused in a feature-integration layer via element-wise addition to combine complementary representations.
The proposed framework is evaluated using the benchmark Lorenz chaotic system under two representative scenarios: (i) autonomous evolution prediction, which assesses the model's ability to replicate complex dynamics over extended horizons without external input~\cite{viehweg2025deterministic,hu2024attractor,cai2024reinforced,wang2024chaotic,yuan2024optoelectronic}; and (ii) inference of unmeasured variables, which reconstructs unobserved states from limited measurements~\cite{lu2017reservoir,feng2022less}. Experimental results demonstrate that the hybrid model consistently achieves superior predictive accuracy and stability across both tasks, confirming its robustness and effectiveness in chaotic system forecasting.

\section{Model Design and Methodology}
\subsection{Architecture}

\begin{figure}
    \centering
    \includegraphics[width=1\columnwidth]{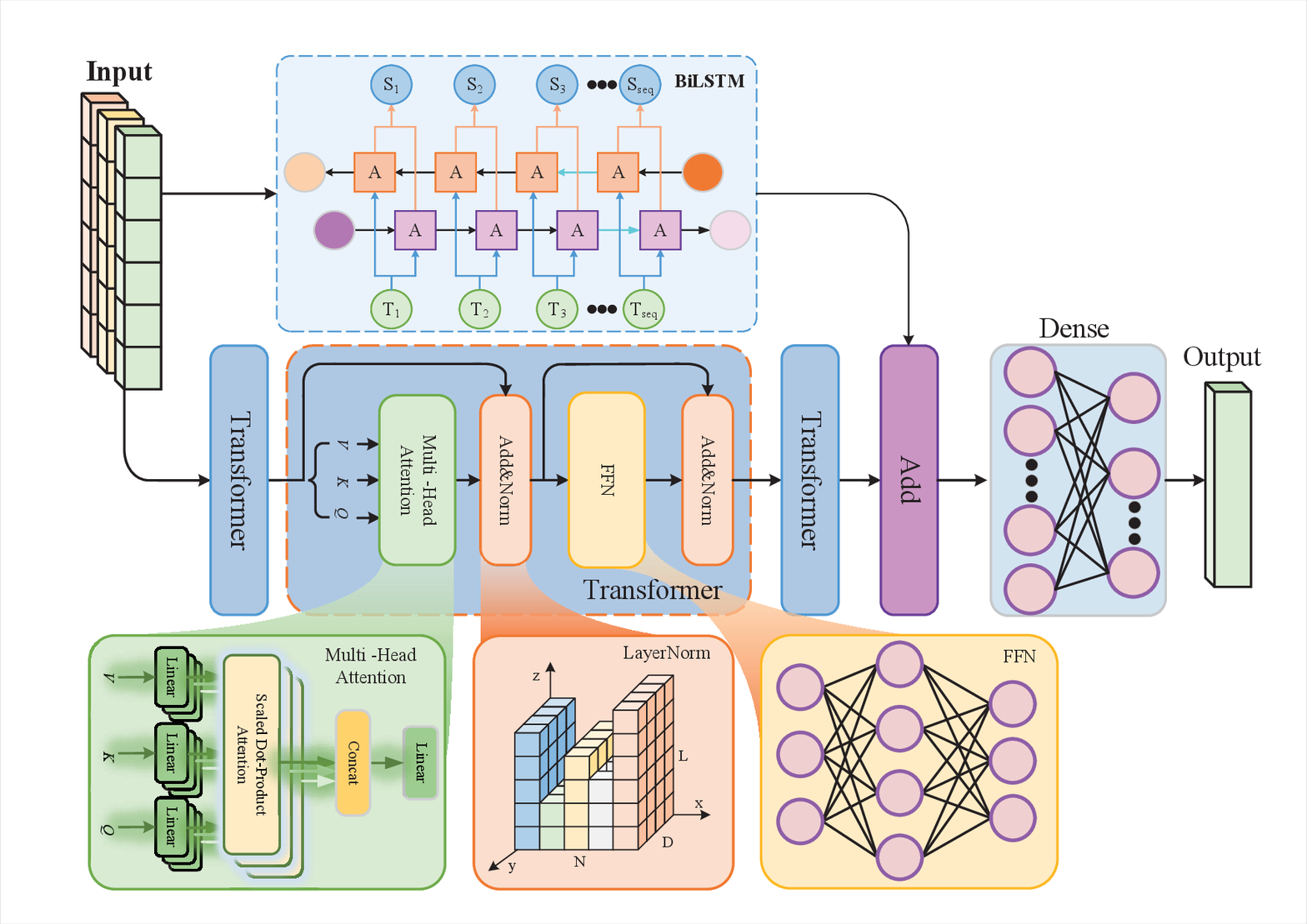}
    \caption{Structure of the proposed dual branch model, comprising a BiLSTM branch and a Transformer branch. The outputs of the BiLSTM and Transformer branches are merged through element-wise addition, and the fused representation is subsequently passed to a dense layer for final prediction.}\label{fig:model}
\end{figure}

The model proposed in this study employs a parallel dual-branch architecture that integrates Transformer and BiLSTM networks. As illustrated in Fig.~\ref{fig:model}, the overall framework comprises three main components: a BiLSTM branch, a Transformer branch, and a feature-fusion module. After standard data preprocessing, the input sequence is simultaneously fed into both branches for feature extraction. The BiLSTM branch is designed to capture local temporal dynamics through bidirectional recurrent processing, whereas the Transformer branch utilizes self-attention mechanisms to model long-range dependencies and global structural correlations.

The feature representations extracted by the two branches are subsequently merged in the feature-fusion module via element-wise addition, producing an integrated representation that combines the complementary strengths of local and global feature modelling. This fused representation is then passed to a fully connected output layer to perform chaotic time series forecasting. The following subsections provide a detailed mathematical formulation of each component, describe the fusion mechanism, and outline the overall training procedure.

\subsection{Methods}

The proposed approach consists of four main stages: (i) data preprocessing, (ii) feature extraction, (iii) feature fusion, and (iv) prediction. The stages are detailed below.

\emph{i. Data Preprocessing}

The input sequence $\{\boldsymbol{x}_1, \boldsymbol{x}_2, \dots, \boldsymbol{x}_T\}$ is first normalized to ensure consistent feature scales across the model inputs. The normalization process is performed as
\begin{equation}
\hat{x}_{t,k} = 2\frac{x_{t,k}-x_{\min,k}}{x_{\max,k}-x_{\min,k}} - 1,\quad k=1,2,\dots,d_x,
\end{equation}
where $x_{t,k}$ denotes the $k$-th feature at time $t$, $x_{\min,k}$ and $x_{\max,k}$ are the minimum and maximum values of the $k$-th feature in the training dataset, $\hat{x}_{t,k}$ is the normalized value, and $d_x$ is the input feature dimension.

\emph{ii. Feature Extraction}

The feature extraction stage consists of two modules: BiLSTM and Transformer, each designed to capture distinct temporal characteristics.

(a) BiLSTM module---The BiLSTM module comprises two parallel LSTM branches that process the sequence in forward and backward directions, thereby capturing bidirectional temporal dependencies \cite{beck2024}. At each time step, the hidden states vector $\boldsymbol{h}_t$ from the two directions are concatenated to form a comprehensive representation of the current features
\begin{equation}
\boldsymbol{h}_t = \left[\overrightarrow{\boldsymbol{h}}_t;\overleftarrow{\boldsymbol{h}}_t\right],
\end{equation}
where $\overrightarrow{\boldsymbol{h}}_t$ and $\overleftarrow{\boldsymbol{h}}_t$ denote the forward and backward hidden states, respectively. The final output of this module is obtained as
\begin{equation}
\boldsymbol{f}_{\rm L} = \boldsymbol{W} \boldsymbol{h}_t + \boldsymbol{b},
\end{equation}
with $\boldsymbol{W}$ and $\boldsymbol{b}$ denoting weight matrix and bias vector.

(b) Transformer module---As illustrated in Fig.~\ref{fig:model}, the Transformer module operates in parallel with the BiLSTM branch to extract long-term dependencies. The normalized input sequence is first linearly projected and then combined with positional encoding, which explicitly injects temporal order information into the feature representation~\cite{ashish2017attention}.
Let $\boldsymbol{z}_t^{(l)} \in \mathbb{R}^{d_m}$ denote the feature vector at time step $t$ in the $l$-th layer, and let
$\boldsymbol{Z}^{(l)} = [\boldsymbol{z}_1^{(l)}, \boldsymbol{z}_2^{(l)}, \dots, \boldsymbol{z}_T^{(l)}]^{\mathrm{T}} \in \mathbb{R}^{T \times d_m}$
represent the feature matrix composed of all time steps, where $l = 0, 1, \dots, L$ denotes the layer index, $L$ is the total number of encoder layers, and $d_m$ is the feature dimension of the Transformer. The initialization process can be expressed as
\begin{equation}
\boldsymbol{z}_t^{(0)} = \boldsymbol{W}_p \boldsymbol{x}_t + \boldsymbol{b}_p + \boldsymbol{E}_t,
\end{equation}
where $\boldsymbol{x}_t$ is the input feature vector at time $t$, $\boldsymbol{W}_p$ and $\boldsymbol{b}_p$ are the projection weight matrix and bias vector, respectively, and $\boldsymbol{E}_t$ denotes the positional encoding vector.

The positional encoding $\boldsymbol{E}_t$ is defined using sine and cosine functions as
\begin{equation}
\begin{aligned}
E_{t, 2k}   &= \sin\left(\frac{t}{C^{2k/d_m}}\right), \\
E_{t, 2k+1} &= \cos\left(\frac{t}{C^{2k/d_m}}\right),
\end{aligned}\quad k = 0, 1, \dots, \frac{d_m}{2}-1,
\end{equation}
where $E_{t,2k}$ and $E_{t,2k+1}$ denote the $2k$-th and $(2k+1)$-th elements of the positional encoding at time step $t$, respectively. Here, $k$ is the feature index, and $C$ is a scaling constant. Following the original Transformer implementation, $C$ is set to 10,000~\cite{ashish2017attention}.

After positional encoding, the sequence is fed into the Transformer encoder modules, which employ self-attention mechanisms to capture bidirectional dependencies and model global correlations within the time series. Specifically, the Transformer utilizes multi-head self-attention to effectively regulate the degree of interaction among different temporal positions. In this study, $L = 3$ stacked Transformer encoder layers are adopted. For the $l$-th encoder layer, the multi-head attention is defined as
\begin{equation}
\begin{cases}
\boldsymbol{Q} = \boldsymbol{Z}^{(l-1)}\boldsymbol{W}_Q^{(l)}, \\
\boldsymbol{K} = \boldsymbol{Z}^{(l-1)}\boldsymbol{W}_K^{(l)}, \\
\boldsymbol{V} = \boldsymbol{Z}^{(l-1)}\boldsymbol{W}_V^{(l)},
\end{cases}
\end{equation}
where $\boldsymbol{Q}$, $\boldsymbol{K}$, and $\boldsymbol{V}$ represent the query, key, and value matrices in the self-attention mechanism, respectively. $\boldsymbol{Z}^{(l-1)}$ denotes the output feature matrix from the $(l-1)$-th layer, while $\boldsymbol{W}_Q^{(l)}$, $\boldsymbol{W}_K^{(l)}$, and $\boldsymbol{W}_V^{(l)}$ are learnable projection weight matrices.

The multi-head attention output at the $l$-th layer is computed as
\begin{equation}
\boldsymbol{M}^{(l)} = \mathrm{Concat}(\mathrm{head}_1,\dots,\mathrm{head}_h)\boldsymbol{W}_O^{(l)},
\end{equation}
where $\boldsymbol{W}_O^{(l)}$ is the output projection matrix. The residual connection and normalization process are then performed as
\begin{equation}
\boldsymbol{\bar{Z}}^{(l)} = \mathrm{LayerNorm}\left(\boldsymbol{Z}^{(l-1)} + \boldsymbol{M}^{(l)}\right),
\end{equation}
followed by the position-wise feed-forward network (FFN), i.e.,
\begin{equation}
\boldsymbol{Z}^{(l)} = \mathrm{LayerNorm}\left(\boldsymbol{\bar{Z}}^{(l)} + \mathrm{FFN}\left(\boldsymbol{\bar{Z}}^{(l)}\right)\right),
\end{equation}
where $\boldsymbol{\bar{Z}}^{(l)}$ and $\boldsymbol{Z}^{(l)}$ denote the intermediate and final output feature matrices after attention and residual operations, respectively, and $\mathrm{FFN}(\cdot)$ represents the position-wise feed-forward network.

Finally, the feature representation of the Transformer at the last time step is expressed as
\begin{equation}
\boldsymbol{f}_{\mathrm{Tr}} = \boldsymbol{Z}_T^{(L)}.
\end{equation}

\emph{iii. Feature Fusion Module}

The proposed model employs element-wise summation in the fusion layer to integrate complementary features extracted from the BiLSTM and Transformer branches. Prior to fusion, the outputs of both branches are linearly projected into a common latent space, whose dimensionality is set to the larger of the two feature dimensions. Specifically, the features produced by the BiLSTM network, $\boldsymbol{f}_{\mathrm{L}}$, and those generated by the Transformer network, $\boldsymbol{f}_{\mathrm{Tr}}$, are mapped to the same dimension and then combined by element-wise addition as
\begin{equation}
\boldsymbol{f} = \boldsymbol{f}_{\mathrm{L}} + \boldsymbol{f}_{\mathrm{Tr}} \in \mathbb{R}^{d_m^{\max}},
\end{equation}
where $d_m^{\max}$ denotes the maximum feature dimension between the BiLSTM and Transformer branches.

\emph{iv. Output Layer}

After feature fusion, the resulting vector is linearly projected into the output space as
\begin{equation}
\boldsymbol{\hat{y}} = \boldsymbol{W}_o \boldsymbol{f} + \boldsymbol{b}_o \in \mathbb{R}^{O_w d_{\mathrm{out}}},
\end{equation}
and reshaped to form the final prediction matrix
\begin{equation}
\boldsymbol{\hat{Y}} = \mathrm{reshape}\left(\boldsymbol{\hat{y}}, (O_w, d_{\mathrm{out}})\right).
\end{equation}
Here, $\boldsymbol{W}_o$ and $\boldsymbol{b}_o$ denote the weight matrix and bias vector of the output projection layer, respectively. $\boldsymbol{\hat{y}}$ represents the flattened predicted output, while $\boldsymbol{\hat{Y}}$ denotes the final prediction matrix, where each row corresponds to a prediction time step and each column to an output dimension. $O_w$ is the prediction window length (i.e., the number of forecast steps), and $d_{\mathrm{out}}$ is the dimensionality of the output variables.

\section{Experiments and analysis of results}

\subsection{Design of Experiments}

The classical Lorenz system is employed as the benchmark to evaluate the model's performance. Its three-dimensional dynamical equations are given by
\begin{subequations}\label{eq:lorenz}
\begin{align}
\dot{x} &= \sigma (y - x), \label{eq:lorenz_a}\\
\dot{y} &= x(\rho - z) - y, \label{eq:lorenz_b}\\
\dot{z} &= xy - \beta z, \label{eq:lorenz_c}
\end{align}
\end{subequations}
where $\sigma$, $\rho$, and $\beta$ are system parameters. In this study, the parameters are set to $\sigma = 10$, $\rho = 28$, and $\beta = 8/3$. Under these parameter settings, the largest Lyapunov exponent (LLE) of the Lorenz system is approximately $0.9056$, indicating strong sensitivity to initial conditions and the presence of typical chaotic behaviour~\cite{cvitanovic_universality_2017}.

To balance simulation accuracy and practical observation conditions, the fourth-order Runge-Kutta (RK4) algorithm is employed with a small integration step of $h=10^{-3}$ to generate high-precision reference trajectories. Based on these trajectories, training data are sampled at larger intervals ($\Delta=60h=0.06$) to emulate realistic scenarios with limited computational resources and low observation frequency. A total of 5000 data points are generated, corresponding to a time span of $t \in [-300, 0]$ for the training dataset. This configuration ensures that the model is trained on a sufficiently long chaotic sequence to capture both short-term fluctuations and long-term system behaviour, thereby enabling a comprehensive evaluation of prediction performance.

\begin{table}
\caption{\label{tab:hyperparams}%
Hyperparameter settings for models.}
\begin{ruledtabular}
\begin{tabular}{llc}
\textbf{Category} & \textbf{Parameter} & \textbf{Value} \\
\hline
\multirow{5}{*}{Data processing}
  & Training data points   & 4000 \\
  & Validate data points   & 500 \\
  & Test data points       & 500 \\
\hline
\multirow{3}{*}{General parameters}
  & Batch size  & 16 \\
  & Sequence length    & 10 \\
  & Prediction length   & 1 \\
\hline
\multirow{1}{*}{BiLSTM}
  & Hidden\_dim  & 256 \\
\hline
\multirow{5}{*}{Transformer}
  & $d_m$   & 64 \\
  & $L$     & 3 \\
  & Trans\_heads      & 8 \\
  & FFN & 256 \\
  & Dropout & 0.1 \\
\end{tabular}
\end{ruledtabular}
\end{table}

The model training follows a supervised learning framework. The input sequence is processed through the dual-branch architecture to extract spatiotemporal representations, and the predicted outputs are obtained via the feature fusion module. The mean squared error (MSE) is employed as the loss function for parameter optimization, defined as
\begin{equation}\label{eq-MSE}
  loss=\frac{1}{N_{\rm trian}}\sum_{i=1}^{N_{\rm trian}}\|\hat{\boldsymbol{x}}_i-\boldsymbol{x}_i\|^2,
\end{equation}
where $\boldsymbol{x}_i$ and $\hat{\boldsymbol{x}}_i$ respectively represent the true and predicted values of the $i$-th sample (here $\boldsymbol{x}_i=[x_i,y_i,z_i]$ is a three-dimensional vector), $N_{\rm trian}$ is the total number of training samples, and $\|\cdot\|$ denotes the Euclidean norm.
The Adam optimizer is used with an initial learning rate of 0.001. A dynamic learning-rate adjustment strategy is applied: if the validation loss does not improve for five consecutive epochs, the learning rate is reduced by half. The maximum number of training epochs is set to 200, and an early-stopping mechanism is introduced to prevent overfitting---training terminates if the validation loss fails to improve for 15 consecutive epochs.

During training, each batch contains 16 samples, and the training set is randomly shuffled before every epoch to enhance model generalization. The model parameters corresponding to the best validation performance are selected for final evaluation, ensuring efficient and stable time series prediction. The detailed experimental settings and hyperparameters are summarized in Table~\ref{tab:hyperparams}.

To comprehensively assess the predictive capability of the proposed model under different observation constraints, two categories of experiments are conducted: (i) autonomous evolution prediction and (ii) inference of unmeasured variables.

\subsection{Experiments}

\subsubsection{Autonomous evolution prediction}

In the autonomous evolution prediction task, the model recursively forecasts future trajectories from historical data~ \cite{viehweg2025deterministic,hu2024attractor,cai2024reinforced,wang2024chaotic,yuan2024optoelectronic}. The process can be formally expressed as
\begin{equation}\label{eq-Self-E}
\boldsymbol{x}_{n+1}=F(\boldsymbol{x}_1,\boldsymbol{x}_2,\dots,\boldsymbol{x}_n),
\end{equation}
where $\boldsymbol{x}_n$ denotes the state vector at the $n$-th time step, and $\boldsymbol{x}_{n+1}$ is the subsequent state predicted by the model $F$ from the preceding $n$ states. To quantify prediction accuracy, the temporal evolution of the normalized root-mean-square error (NRMSE)~\cite{2021-chaos} is monitored, defined as
\begin{equation}
E_j = \frac{\|\hat{\boldsymbol{x}}_j - \boldsymbol{x}_j\|}{\sqrt{\frac{1}{j}\sum_{i=1}^{j}\|\boldsymbol{x}_i\|^2}}.
\end{equation}
Based on the NRMSE, the valid prediction time (VPT) is introduced as the time $t_{\mathrm{VPT}} = j\Delta$ at which the NRMSE first exceeds a preset threshold $E_{\mathrm{c}}$, i.e.,
\begin{equation}
t_{\mathrm{VPT}} = \min \left\{ t = j\Delta ,|, E_j > E_{\mathrm{c}} \right\}.
\end{equation}
Following Refs.~\cite{2021-chaos,chepuri2024hybridizing}, the threshold is set to $E_{\mathrm{c}} = 0.9$. To express prediction horizons in a physically meaningful manner, the VPT is normalized by the Lyapunov time, defined as the inverse of the system's LLE.

\begin{figure}[t]
    \centering
    \includegraphics[width=1\columnwidth]{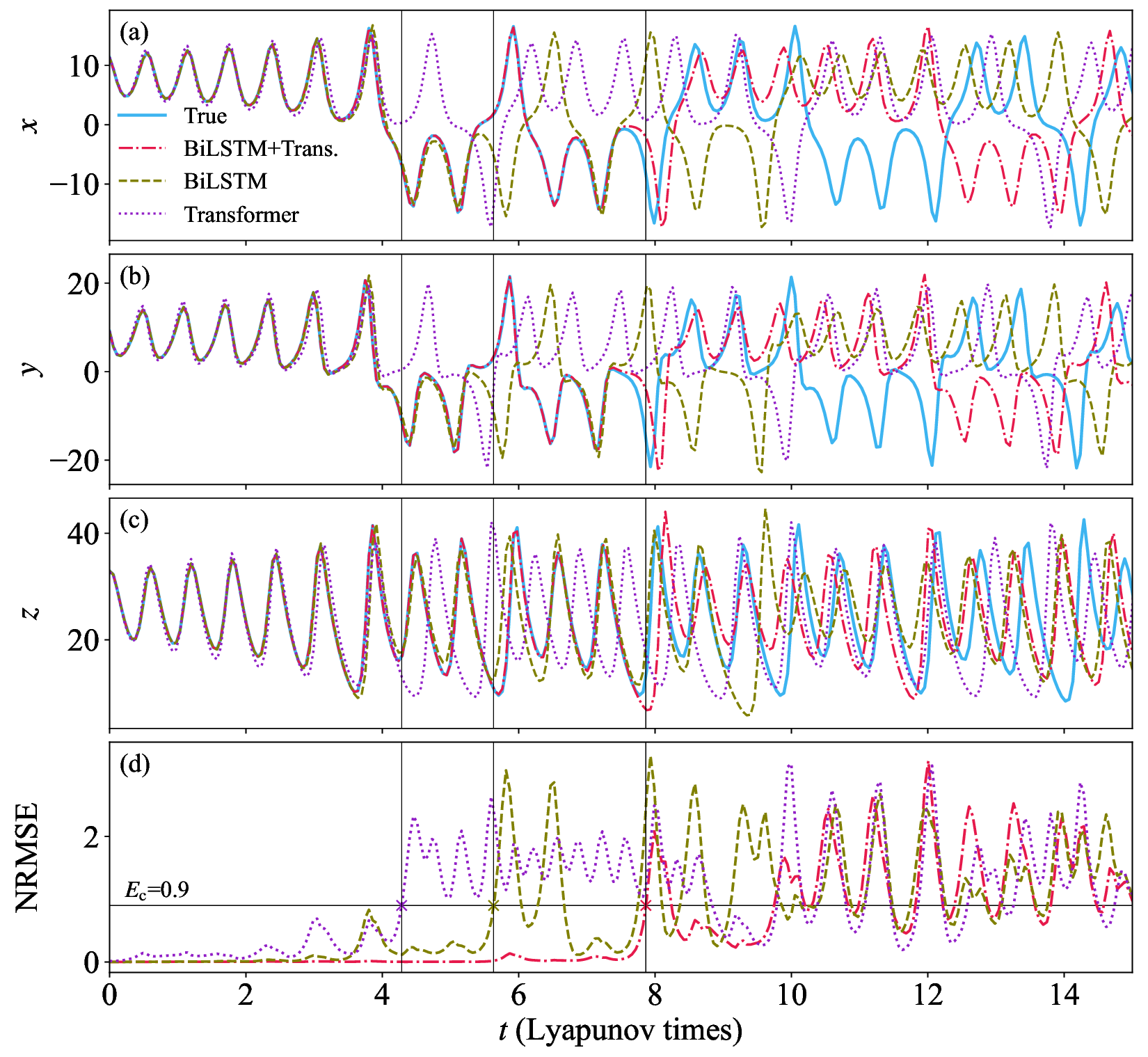}
    \caption{The prediction results of different models on the autonomous evolution of the Lorenz system. (a)-(c) correspond respectively to the results of state variables $x$, $y$, and $z$. (d) Evolution of NRMSE of the three models. The intersection point of the horizontal solid line and the curve (indicated by the cross) determines the VPT. All panels share the legend.}
    \label{fig2:enter-label}
\end{figure}

\begin{figure}[t]
    \centering
    \includegraphics[width=1\columnwidth]{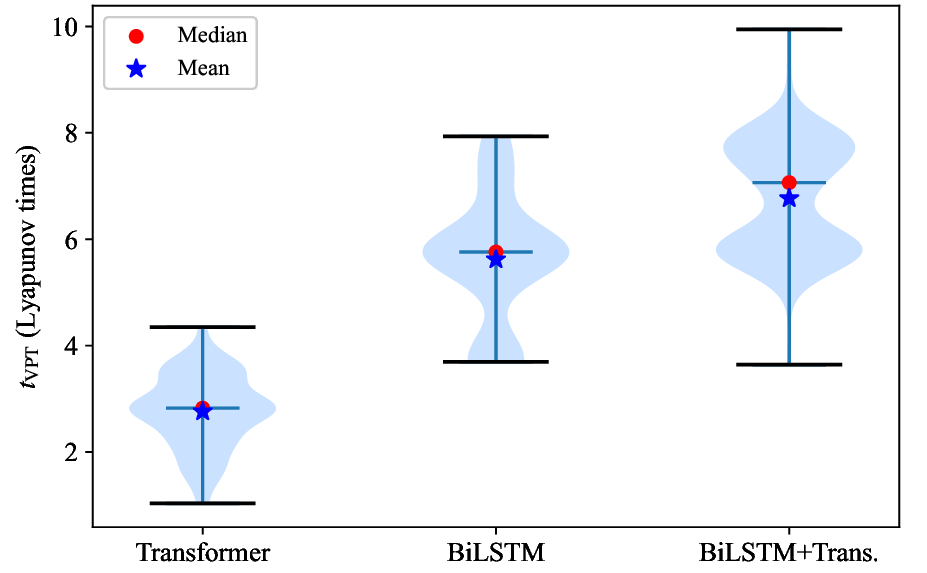}\\
    \caption{The violin plot shows the distribution of $t_{\rm VPT}$ given by different models under 100 random initializations, where the red dots and blue stars correspond to the median and mean, respectively.}
    \label{fig3:enter-label}
\end{figure}

Figures~\ref{fig2:enter-label}(a)-(c) display the time evolution of the true and predicted trajectories for the state variables $x$, $y$, and $z$, respectively, while Fig.~\ref{fig2:enter-label}(d) presents the corresponding NRMSE curves. The hybrid BiLSTM-Transformer model achieves significantly higher prediction accuracy than either individual architecture, as indicated by the slower error growth in Fig.~\ref{fig2:enter-label}(d). Among the baseline models, BiLSTM performs second best, while the Transformer alone exhibits a faster divergence from the ground truth.

To further evaluate the robustness of the models, 100 independent experiments were conducted for each architecture under different random initializations. The median VPT was adopted as the evaluation metric to mitigate the influence of statistical outliers. Figure~\ref{fig3:enter-label} presents the distribution of VPT obtained from these experiments. The proposed hybrid model yields a median VPT of 7.06 Lyapunov times, substantially exceeding those of the BiLSTM (5.76) and Transformer (2.83) models. These results clearly demonstrate the superior predictive stability and robustness of the proposed BiLSTM-Transformer hybrid architecture in the autonomous evolution prediction task.

\begin{figure}[t]
    \centering
    \includegraphics[width=1\columnwidth]{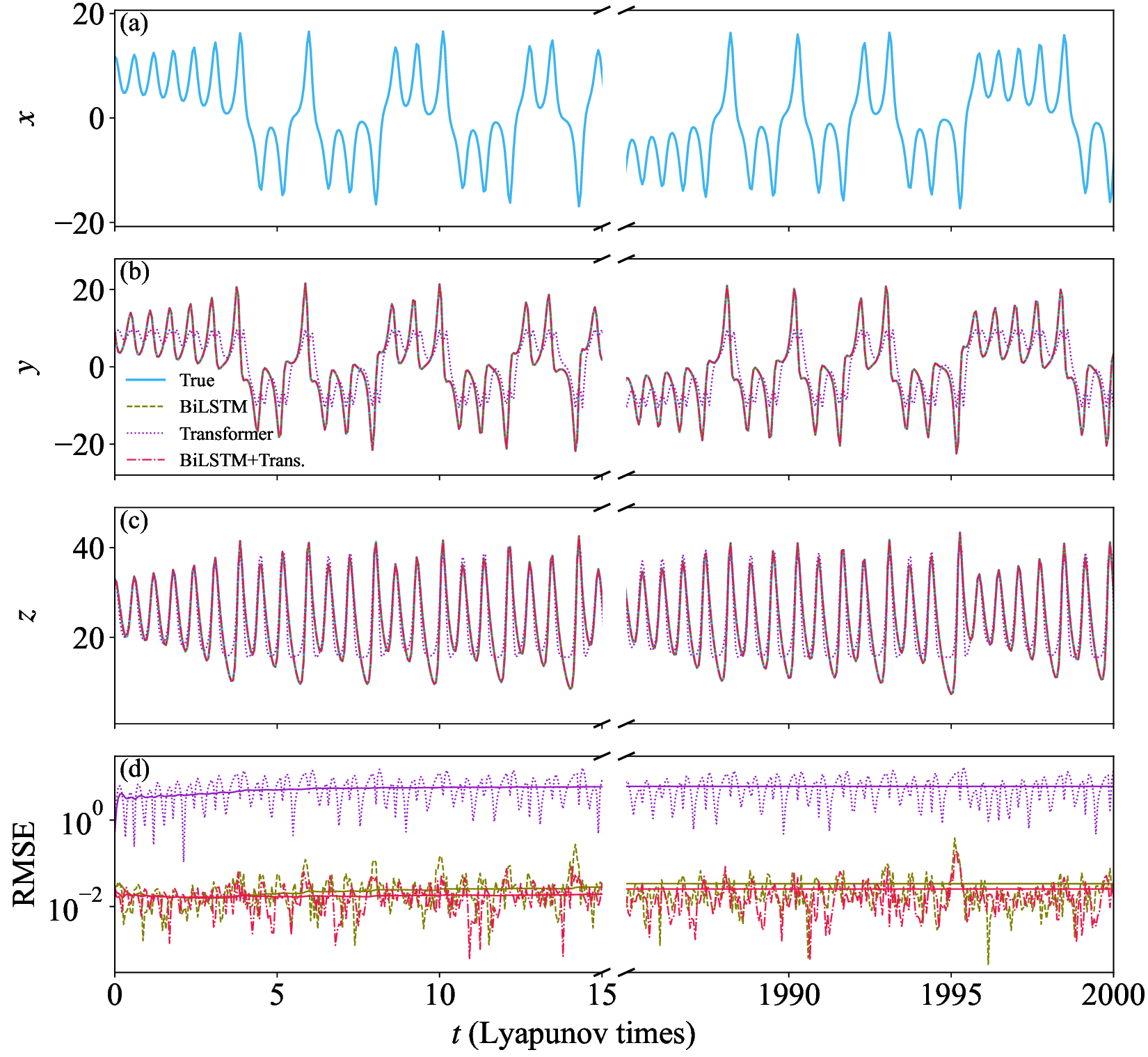}
    \caption{Inferring results of $y$ (b) and $z$ (c) using $x$ (a) as the observed variable, i.e., $\boldsymbol{m}=x$ and $\boldsymbol{u}=[y,z]$ in Eq.~(\ref{eq-infer-E}). (d) The evolution of RMSE corresponding to different models, where the solid line represents their cumulative average. Panels (b)-(d) share the legend.}
    \label{fig4:enter-label}
\end{figure}

\subsubsection{ Inference of unmeasured variables}

In the task of inferring unmeasured system states from partially observable variables, the model receives as input the time-delay embedding vectors constructed from the observed variables and outputs the corresponding estimates of the unmeasured states. This process can be mathematically expressed as
\begin{equation}\label{eq-infer-E}
  \hat{\boldsymbol{u}}_{n+1}=F(\boldsymbol{m}_1,\boldsymbol{m}_2,\dots,\boldsymbol{m}_n),
\end{equation}
where $\boldsymbol{m}_n$ denotes the measured state vector of the system at the $n$-th time step, and $\hat{\boldsymbol{u}}_{n+1}$ represents the predicted unmeasured state vector at $(n+1)$.
The inference accuracy is quantitatively evaluated using the RMSE, defined as
\begin{equation}
\mathrm{RMSE}(t) = \|\hat{\boldsymbol{u}}(t) - \boldsymbol{u}(t)\|,
\end{equation}
where $\hat{\boldsymbol{u}}(t)$ and $\boldsymbol{u}(t)$ denote the inferred and true values of the unmeasured variables at time step $t$, respectively.

Using the Lorenz system as an illustrative example (i.e., on the same dataset), the variable $x$ is selected as the observed input to infer the remaining state variables $y$ and $z$; that is, $\boldsymbol{m}=x$ and $\boldsymbol{u}=[y,z]$ in Eq.~(\ref{eq-infer-E}). The corresponding results are shown in Fig.~\ref{fig4:enter-label}. Figure~\ref{fig4:enter-label}(a) displays the observed time series of $x$, while Figs.~\ref{fig4:enter-label}(b) and (c) compare the true (solid lines) and predicted trajectories of $y$ and $z$ obtained by different models, with distinct line styles indicating model types.
As evident from Figs.~\ref{fig4:enter-label}(b)-(c), the BiLSTM and the proposed hybrid model produce trajectories that almost coincide with the ground truth, whereas the Transformer model exhibits significant deviations.

\begin{figure}[t]
    \centering
    \includegraphics[width=1\columnwidth]{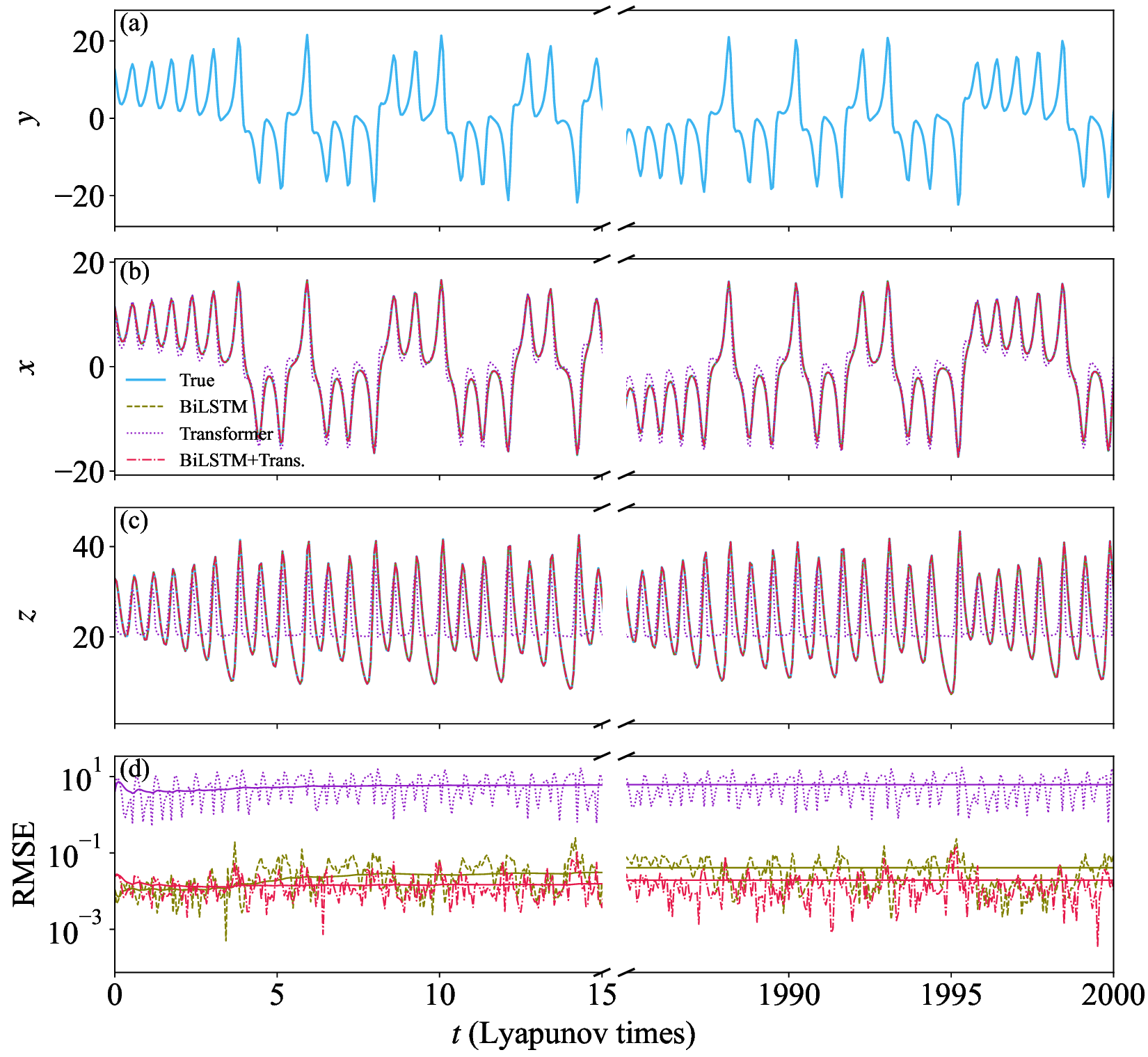}
    \caption{Inferring results of $x$ (b) and $z$ (c) using $y$ (a) as the observed variable. (d) The evolution of RMSE and $\overline{\rm RMSE}$.}
    \label{fig5:enter-label}
\end{figure}
\begin{figure}
    \centering
    \includegraphics[width=1\columnwidth]{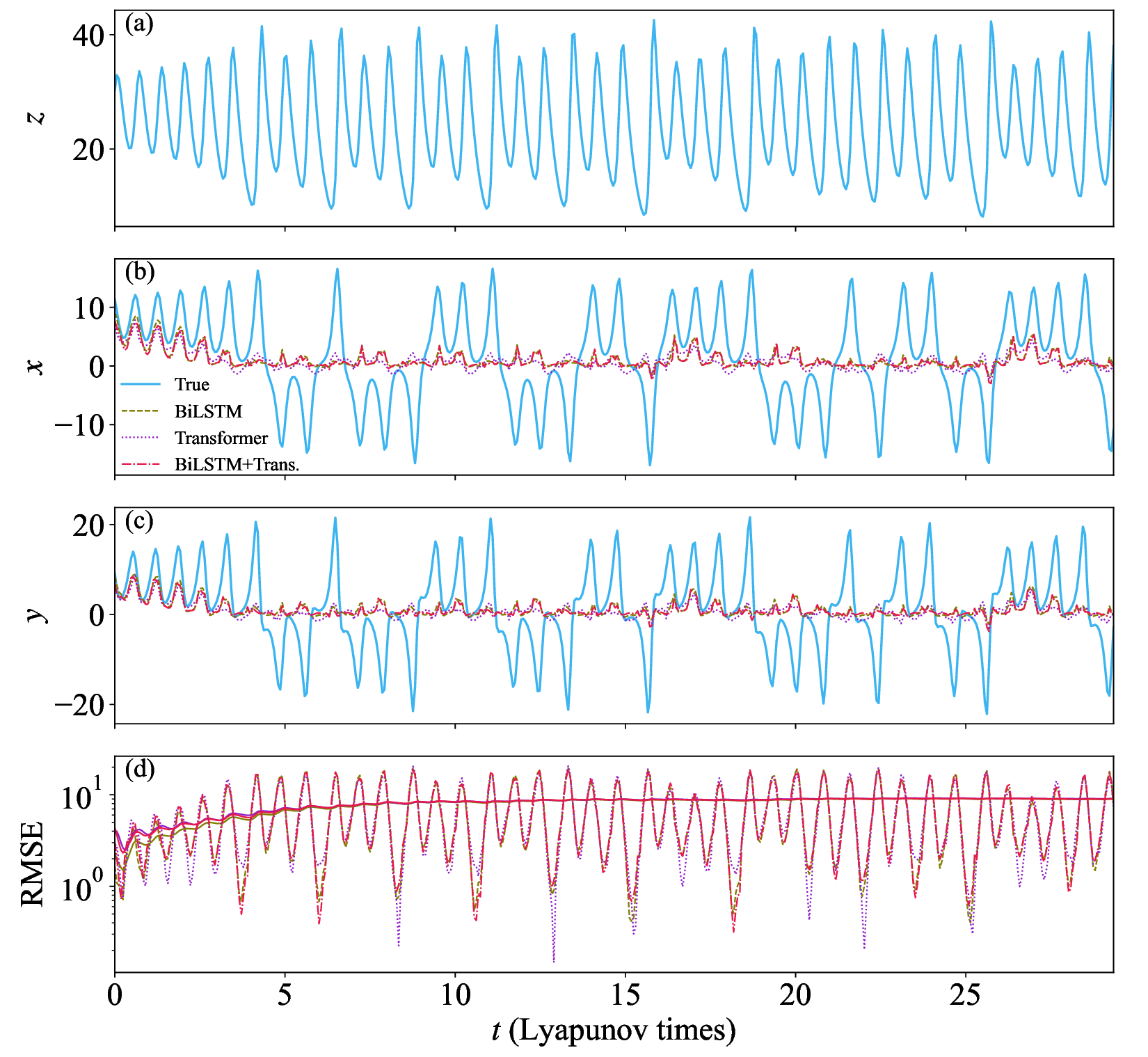}
    \caption{Inferring results of $x$ (b) and $y$ (c) using $z$ (a) as the observed variable. (d) The evolution of RMSE and $\overline{\rm RMSE}$.}
    \label{fig6:enter-label}
\end{figure}

Figure~\ref{fig4:enter-label}(d) presents the temporal evolution of RMSE for the three models. The RMSE of both the BiLSTM (dashed line) and the hybrid model (dash-dotted line) stabilizes around the $10^{-2}$ level, while the hybrid model achieves the lowest cumulative average RMSE, confirming its superior inference performance. The cumulative average RMSE is given by
\begin{equation}
\overline{\mathrm{RMSE}}(t) = \frac{1}{t} \sum_{\tau=1}^{t} \|\hat{\boldsymbol{u}}(\tau) - \boldsymbol{u}(\tau)\|.
\end{equation}
As shown in Fig.~\ref{fig4:enter-label}(d), $\overline{\mathrm{RMSE}}(t)$ gradually converges to a constant value, indicating that the unmeasured state variables can be stably inferred over arbitrarily long horizons using only time-delay embeddings of partial observations. This apparent suppression of exponential error divergence does not signify a breakdown of chaos but rather results from the model's ability to learn the intrinsic coupling among state variables. The continuous access to true observations dynamically corrects the predicted trajectories, effectively preventing cumulative error growth.

\begin{figure}[t]
    \centering
    \includegraphics[width=1\columnwidth]{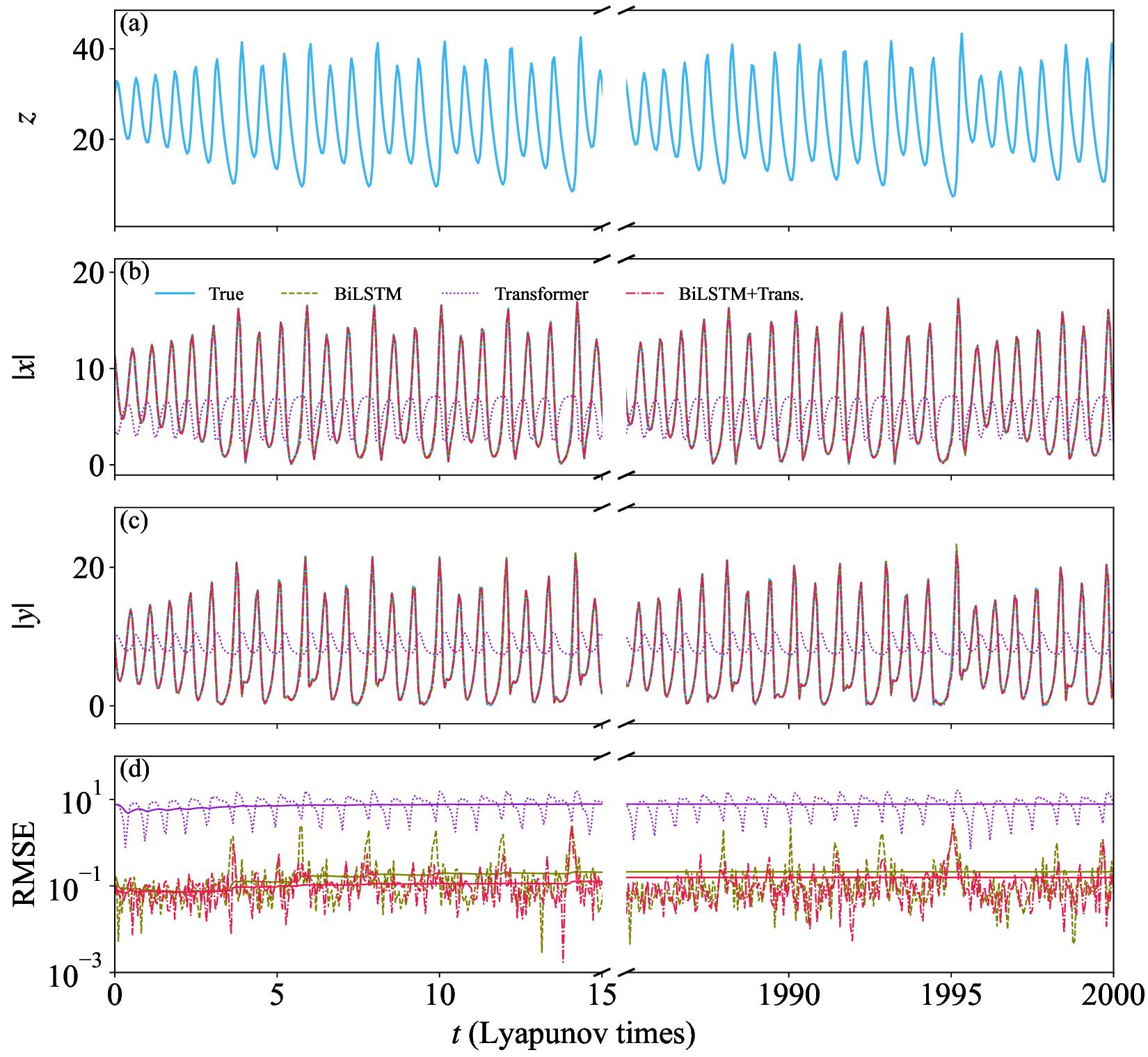}
    \caption{Inferring results of $|x|$ (b) and $|y|$ (c) using $z$ (a) as the observed variable, i.e., $\boldsymbol{m}=z$ and $\boldsymbol{u}=[|x|,|y|]$ in Eq.~(\ref{eq-infer-E}). (d) The evolution of RMSE and $\overline{\rm RMSE}$.}
    \label{fig7:enter-label}
\end{figure}

Figure~\ref{fig5:enter-label} presents the results when $y$ serves as the observed variable, with $x$ and $z$ inferred from its time-delay embedding. The qualitative behavior remains consistent with that in Fig.~\ref{fig4:enter-label}.
In contrast, when $z$ is the sole observed variable, none of the models can accurately reconstruct the evolution of $x$ and $y$, as shown in Fig.~\ref{fig6:enter-label}. This failure originates from the intrinsic symmetry of the Lorenz system: when only $z$ is observed, the states $(x, y)$ and $(-x, -y)$ become indistinguishable, leading to an inherent sign ambiguity~\cite{lu2017reservoir}. Consequently, the inference problem becomes ill-posed, and accurate prediction of $x$ and $y$ from $z$ alone is fundamentally infeasible.

To eliminate this indistinguishability, we use $z$ to infer the absolute values of $x$ and $y$. The corresponding results, shown in Fig.~\ref{fig7:enter-label}, exhibit qualitative behavior consistent with that in Figs.~\ref{fig4:enter-label} and \ref{fig5:enter-label}. Across all cases, the proposed BiLSTM-Transformer hybrid model consistently achieves the best overall performance, demonstrating strong generalization and stability in the inference of unmeasured variables.

\section{Summary and Discussions}

In summary, we have proposed a hybrid time-series prediction framework featuring a parallel dual-branch architecture that integrates Transformer and BiLSTM modules. The Transformer branch primarily captures global dependencies within chaotic sequences through self-attention, while the BiLSTM branch focuses on local temporal dynamics via bidirectional recurrence. The complementary representations extracted by the two branches are effectively integrated through element-wise feature fusion, substantially enhancing the model's ability to characterize and predict complex nonlinear dynamical systems. Using the Lorenz system as a representative benchmark, the proposed model was systematically evaluated on two canonical tasks---autonomous evolution prediction and inference of unmeasured variables. The results consistently show that the hybrid model delivers the highest overall performance, while the BiLSTM surpasses the Transformer among the single-branch networks.

Furthermore, the experiments reveal a fundamental difference between the two tasks. The inference of unmeasured variables is intrinsically easier than autonomous evolution prediction, as real-time observations provide continuous feedback that dynamically corrects the system trajectories, allowing stable long-term estimation. In contrast, autonomous evolution suffers from error accumulation and the exponential divergence characteristic of chaotic dynamics, which inherently restrict the prediction horizon to a finite number of Lyapunov times.

The demonstrated capability to infer unmeasured variables carries significant implications for practical applications. It enables the estimation of difficult-to-measure or costly variables using only accessible observations, which is particularly valuable in engineering diagnostics, environmental and health monitoring,  investigation of coupling relationships among physiological or system indicators, and complex-system control. Beyond these domains, the hybrid framework provides a general and extensible methodology for data-driven modeling of nonlinear dynamical systems, offering a promising route toward accurate, interpretable, and computationally efficient prediction of chaotic behaviors.

\begin{acknowledgments}

This work was supported by the National Natural Science Foundation of China (Grants Nos. 12465010, 12247106, 12575040, 12575042, 12005156, 12247101). W. Fu also acknowledges support from the Youth Talent (Team) Project of Gansu Province (No. 2024QNTD54), the Long-yuan Youth Talents Project of Gansu Province, the Fei-tian Scholars Project of Gansu Province, the Leading Talent Project of Tianshui City, the Innovation Fund from the Department of Education of Gansu Province (Grant No.~2023A-106), the Open Project Program of Key Laboratory of Atomic and Molecular Physics $\&$ Functional Material of Gansu Province (Grant No. 6016-202404), and the Project of Open Competition for the Best Candidates from Department of Education of Gansu Province (Grant No. 2021jyjbgs-06).

\end{acknowledgments}

\bibliography{Refs}

\begin{thebibliography}{51}%
\makeatletter
\providecommand \@ifxundefined [1]{%
 \@ifx{#1\undefined}
}%
\providecommand \@ifnum [1]{%
 \ifnum #1\expandafter \@firstoftwo
 \else \expandafter \@secondoftwo
 \fi
}%
\providecommand \@ifx [1]{%
 \ifx #1\expandafter \@firstoftwo
 \else \expandafter \@secondoftwo
 \fi
}%
\providecommand \natexlab [1]{#1}%
\providecommand \enquote  [1]{``#1''}%
\providecommand \bibnamefont  [1]{#1}%
\providecommand \bibfnamefont [1]{#1}%
\providecommand \citenamefont [1]{#1}%
\providecommand \href@noop [0]{\@secondoftwo}%
\providecommand \href [0]{\begingroup \@sanitize@url \@href}%
\providecommand \@href[1]{\@@startlink{#1}\@@href}%
\providecommand \@@href[1]{\endgroup#1\@@endlink}%
\providecommand \@sanitize@url [0]{\catcode `\\12\catcode `\$12\catcode
  `\&12\catcode `\#12\catcode `\^12\catcode `\_12\catcode `\%12\relax}%
\providecommand \@@startlink[1]{}%
\providecommand \@@endlink[0]{}%
\providecommand \url  [0]{\begingroup\@sanitize@url \@url }%
\providecommand \@url [1]{\endgroup\@href {#1}{\urlprefix }}%
\providecommand \urlprefix  [0]{URL }%
\providecommand \Eprint [0]{\href }%
\providecommand \doibase [0]{https://doi.org/}%
\providecommand \selectlanguage [0]{\@gobble}%
\providecommand \bibinfo  [0]{\@secondoftwo}%
\providecommand \bibfield  [0]{\@secondoftwo}%
\providecommand \translation [1]{[#1]}%
\providecommand \BibitemOpen [0]{}%
\providecommand \bibitemStop [0]{}%
\providecommand \bibitemNoStop [0]{.\EOS\space}%
\providecommand \EOS [0]{\spacefactor3000\relax}%
\providecommand \BibitemShut  [1]{\csname bibitem#1\endcsname}%
\let\auto@bib@innerbib\@empty
\bibitem [{\citenamefont
  {Cvitanovi\'{c}}(1989)}]{cvitanovic_universality_2017}%
  \BibitemOpen
  \bibinfo {editor} {\bibfnamefont {P.}~\bibnamefont {Cvitanovi\'{c}}},\ ed.,\
  \href@noop {} {\emph {\bibinfo {title} {Universality in Chaos}}},\ \bibinfo
  {edition} {2nd}\ ed.\ (\bibinfo  {publisher} {Taylor \& Francis Group, LLC},\
  \bibinfo {address} {New York},\ \bibinfo {year} {1989})\BibitemShut {NoStop}%
\bibitem [{\citenamefont {Ratas}\ and\ \citenamefont
  {Pyragas}(2024)}]{ratas2024application}%
  \BibitemOpen
  \bibfield  {author} {\bibinfo {author} {\bibfnamefont {I.}~\bibnamefont
  {Ratas}}\ and\ \bibinfo {author} {\bibfnamefont {K.}~\bibnamefont
  {Pyragas}},\ }\bibfield  {title} {\bibinfo {title} {Application of
  next-generation reservoir computing for predicting chaotic systems from
  partial observations},\ }\href {https://doi.org/10.1103/PhysRevE.109.064215}
  {\bibfield  {journal} {\bibinfo  {journal} {Phys. Rev. E}\ }\textbf {\bibinfo
  {volume} {109}},\ \bibinfo {pages} {064215} (\bibinfo {year}
  {2024})}\BibitemShut {NoStop}%
\bibitem [{\citenamefont {Rand}\ and\ \citenamefont
  {Young}(1981)}]{takens1981dynamical}%
  \BibitemOpen
  \bibinfo {editor} {\bibfnamefont {D.}~\bibnamefont {Rand}}\ and\ \bibinfo
  {editor} {\bibfnamefont {L.-S.}\ \bibnamefont {Young}},\ eds.,\ \href
  {https://doi.org/10.1007/BFb0091903} {\emph {\bibinfo {title} {Dynamical
  Systems and Turbulence}}},\ \bibinfo {series} {Lecture Notes in Mathematics},
  Vol.\ \bibinfo {volume} {898}\ (\bibinfo  {publisher} {Springer Berlin,
  Heidelberg},\ \bibinfo {year} {1981})\BibitemShut {NoStop}%
\bibitem [{\citenamefont {Packard}\ \emph {et~al.}(1980)\citenamefont
  {Packard}, \citenamefont {Crutchfield}, \citenamefont {Farmer},\ and\
  \citenamefont {Shaw}}]{packard1980geometry}%
  \BibitemOpen
  \bibfield  {author} {\bibinfo {author} {\bibfnamefont {N.~H.}\ \bibnamefont
  {Packard}}, \bibinfo {author} {\bibfnamefont {J.~P.}\ \bibnamefont
  {Crutchfield}}, \bibinfo {author} {\bibfnamefont {J.~D.}\ \bibnamefont
  {Farmer}},\ and\ \bibinfo {author} {\bibfnamefont {R.~S.}\ \bibnamefont
  {Shaw}},\ }\bibfield  {title} {\bibinfo {title} {Geometry from a time
  series},\ }\href {https://doi.org/10.1103/PhysRevLett.45.712} {\bibfield
  {journal} {\bibinfo  {journal} {Phys. Rev. Lett.}\ }\textbf {\bibinfo
  {volume} {45}},\ \bibinfo {pages} {712} (\bibinfo {year} {1980})}\BibitemShut
  {NoStop}%
\bibitem [{\citenamefont {Hart}(2023)}]{hart2023estimating}%
  \BibitemOpen
  \bibfield  {author} {\bibinfo {author} {\bibfnamefont {J.~D.}\ \bibnamefont
  {Hart}},\ }\bibfield  {title} {\bibinfo {title} {Estimating the master
  stability function from the time series of one oscillator via reservoir
  computing},\ }\href {https://doi.org/10.1103/PhysRevE.108.L032201} {\bibfield
   {journal} {\bibinfo  {journal} {Phys. Rev. E}\ }\textbf {\bibinfo {volume}
  {108}},\ \bibinfo {pages} {L032201} (\bibinfo {year} {2023})}\BibitemShut
  {NoStop}%
\bibitem [{\citenamefont {Wei}(2022)}]{wei2022reservoir}%
  \BibitemOpen
  \bibfield  {author} {\bibinfo {author} {\bibfnamefont {Z.}~\bibnamefont
  {Wei}},\ }\bibfield  {title} {\bibinfo {title} {Reservoir computing with 2d
  materials},\ }\href {https://doi.org/10.1038/s41928-022-00872-1} {\bibfield
  {journal} {\bibinfo  {journal} {Nature Electronics}\ }\textbf {\bibinfo
  {volume} {5}},\ \bibinfo {pages} {715} (\bibinfo {year} {2022})}\BibitemShut
  {NoStop}%
\bibitem [{\citenamefont {Moon}\ \emph {et~al.}(2019)\citenamefont {Moon},
  \citenamefont {Ma}, \citenamefont {Shin}, \citenamefont {Cai}, \citenamefont
  {Du}, \citenamefont {Lee},\ and\ \citenamefont {Lu}}]{moon2019temporal}%
  \BibitemOpen
  \bibfield  {author} {\bibinfo {author} {\bibfnamefont {J.}~\bibnamefont
  {Moon}}, \bibinfo {author} {\bibfnamefont {W.}~\bibnamefont {Ma}}, \bibinfo
  {author} {\bibfnamefont {J.~H.}\ \bibnamefont {Shin}}, \bibinfo {author}
  {\bibfnamefont {F.}~\bibnamefont {Cai}}, \bibinfo {author} {\bibfnamefont
  {C.}~\bibnamefont {Du}}, \bibinfo {author} {\bibfnamefont {S.~H.}\
  \bibnamefont {Lee}},\ and\ \bibinfo {author} {\bibfnamefont {W.~D.}\
  \bibnamefont {Lu}},\ }\bibfield  {title} {\bibinfo {title} {Temporal data
  classification and forecasting using a memristor-based reservoir computing
  system},\ }\href {https://doi.org/10.1038/s41928-019-0313-3} {\bibfield
  {journal} {\bibinfo  {journal} {Nature Electronics}\ }\textbf {\bibinfo
  {volume} {2}},\ \bibinfo {pages} {480} (\bibinfo {year} {2019})}\BibitemShut
  {NoStop}%
\bibitem [{\citenamefont {Zhong}\ \emph {et~al.}(2022)\citenamefont {Zhong},
  \citenamefont {Tang}, \citenamefont {Li}, \citenamefont {Liang},
  \citenamefont {Liu}, \citenamefont {Li}, \citenamefont {Xi}, \citenamefont
  {Yao}, \citenamefont {Hao}, \citenamefont {Gao}, \citenamefont {Qian},\ and\
  \citenamefont {Wu}}]{zhong2022memristor}%
  \BibitemOpen
  \bibfield  {author} {\bibinfo {author} {\bibfnamefont {Y.}~\bibnamefont
  {Zhong}}, \bibinfo {author} {\bibfnamefont {J.}~\bibnamefont {Tang}},
  \bibinfo {author} {\bibfnamefont {X.}~\bibnamefont {Li}}, \bibinfo {author}
  {\bibfnamefont {X.}~\bibnamefont {Liang}}, \bibinfo {author} {\bibfnamefont
  {Z.}~\bibnamefont {Liu}}, \bibinfo {author} {\bibfnamefont {Y.}~\bibnamefont
  {Li}}, \bibinfo {author} {\bibfnamefont {Y.}~\bibnamefont {Xi}}, \bibinfo
  {author} {\bibfnamefont {P.}~\bibnamefont {Yao}}, \bibinfo {author}
  {\bibfnamefont {Z.}~\bibnamefont {Hao}}, \bibinfo {author} {\bibfnamefont
  {B.}~\bibnamefont {Gao}}, \bibinfo {author} {\bibfnamefont {H.}~\bibnamefont
  {Qian}},\ and\ \bibinfo {author} {\bibfnamefont {H.}~\bibnamefont {Wu}},\
  }\bibfield  {title} {\bibinfo {title} {A memristor-based analogue reservoir
  computing system for real-time and power-efficient signal processing},\
  }\href {https://doi.org/10.1038/s41928-022-00838-3} {\bibfield  {journal}
  {\bibinfo  {journal} {Nature Electronics}\ }\textbf {\bibinfo {volume} {5}},\
  \bibinfo {pages} {672} (\bibinfo {year} {2022})}\BibitemShut {NoStop}%
\bibitem [{\citenamefont {Zhong}\ \emph {et~al.}(2021)\citenamefont {Zhong},
  \citenamefont {Tang}, \citenamefont {Li}, \citenamefont {Gao}, \citenamefont
  {Qian},\ and\ \citenamefont {Wu}}]{zhong2021dynamic}%
  \BibitemOpen
  \bibfield  {author} {\bibinfo {author} {\bibfnamefont {Y.}~\bibnamefont
  {Zhong}}, \bibinfo {author} {\bibfnamefont {J.}~\bibnamefont {Tang}},
  \bibinfo {author} {\bibfnamefont {X.}~\bibnamefont {Li}}, \bibinfo {author}
  {\bibfnamefont {B.}~\bibnamefont {Gao}}, \bibinfo {author} {\bibfnamefont
  {H.}~\bibnamefont {Qian}},\ and\ \bibinfo {author} {\bibfnamefont
  {H.}~\bibnamefont {Wu}},\ }\bibfield  {title} {\bibinfo {title} {Dynamic
  memristor-based reservoir computing for high-efficiency temporal signal
  processing},\ }\href {https://doi.org/10.1038/s41467-020-20692-1} {\bibfield
  {journal} {\bibinfo  {journal} {Nat. Commun.}\ }\textbf {\bibinfo {volume}
  {12}},\ \bibinfo {pages} {408} (\bibinfo {year} {2021})}\BibitemShut
  {NoStop}%
\bibitem [{\citenamefont {Chen}\ \emph {et~al.}(2023)\citenamefont {Chen},
  \citenamefont {Li}, \citenamefont {Fan}, \citenamefont {Dong}, \citenamefont
  {Chen}, \citenamefont {Qin}, \citenamefont {Zeng}, \citenamefont {Lu},
  \citenamefont {Zhou}, \citenamefont {Gao},\ and\ \citenamefont
  {Liu}}]{chen2023all}%
  \BibitemOpen
  \bibfield  {author} {\bibinfo {author} {\bibfnamefont {Z.}~\bibnamefont
  {Chen}}, \bibinfo {author} {\bibfnamefont {W.}~\bibnamefont {Li}}, \bibinfo
  {author} {\bibfnamefont {Z.}~\bibnamefont {Fan}}, \bibinfo {author}
  {\bibfnamefont {S.}~\bibnamefont {Dong}}, \bibinfo {author} {\bibfnamefont
  {Y.}~\bibnamefont {Chen}}, \bibinfo {author} {\bibfnamefont {M.}~\bibnamefont
  {Qin}}, \bibinfo {author} {\bibfnamefont {M.}~\bibnamefont {Zeng}}, \bibinfo
  {author} {\bibfnamefont {X.}~\bibnamefont {Lu}}, \bibinfo {author}
  {\bibfnamefont {G.}~\bibnamefont {Zhou}}, \bibinfo {author} {\bibfnamefont
  {X.}~\bibnamefont {Gao}},\ and\ \bibinfo {author} {\bibfnamefont {J.-M.}\
  \bibnamefont {Liu}},\ }\bibfield  {title} {\bibinfo {title}
  {All-ferroelectric implementation of reservoir computing},\ }\href
  {https://doi.org/10.1038/s41467-023-39371-y} {\bibfield  {journal} {\bibinfo
  {journal} {Nat. Commun.}\ }\textbf {\bibinfo {volume} {14}},\ \bibinfo
  {pages} {3585} (\bibinfo {year} {2023})}\BibitemShut {NoStop}%
\bibitem [{\citenamefont {Brunner}\ \emph {et~al.}(2013)\citenamefont
  {Brunner}, \citenamefont {Soriano}, \citenamefont {Mirasso},\ and\
  \citenamefont {Fischer}}]{brunner2013parallel}%
  \BibitemOpen
  \bibfield  {author} {\bibinfo {author} {\bibfnamefont {D.}~\bibnamefont
  {Brunner}}, \bibinfo {author} {\bibfnamefont {M.~C.}\ \bibnamefont
  {Soriano}}, \bibinfo {author} {\bibfnamefont {C.~R.}\ \bibnamefont
  {Mirasso}},\ and\ \bibinfo {author} {\bibfnamefont {I.}~\bibnamefont
  {Fischer}},\ }\bibfield  {title} {\bibinfo {title} {Parallel photonic
  information processing at gigabyte per second data rates using transient
  states},\ }\href {https://doi.org/10.1038/ncomms2368} {\bibfield  {journal}
  {\bibinfo  {journal} {Nat. Commun.}\ }\textbf {\bibinfo {volume} {4}},\
  \bibinfo {pages} {1364} (\bibinfo {year} {2013})}\BibitemShut {NoStop}%
\bibitem [{\citenamefont {Kan}\ \emph {et~al.}(2022)\citenamefont {Kan},
  \citenamefont {Nakajima}, \citenamefont {Asai},\ and\ \citenamefont
  {Akai-Kasaya}}]{kan2022physical}%
  \BibitemOpen
  \bibfield  {author} {\bibinfo {author} {\bibfnamefont {S.}~\bibnamefont
  {Kan}}, \bibinfo {author} {\bibfnamefont {K.}~\bibnamefont {Nakajima}},
  \bibinfo {author} {\bibfnamefont {T.}~\bibnamefont {Asai}},\ and\ \bibinfo
  {author} {\bibfnamefont {M.}~\bibnamefont {Akai-Kasaya}},\ }\bibfield
  {title} {\bibinfo {title} {Physical implementation of reservoir computing
  through electrochemical reaction},\ }\href
  {https://doi.org/https://doi.org/10.1002/advs.202104076} {\bibfield
  {journal} {\bibinfo  {journal} {Adv. Sci.}\ }\textbf {\bibinfo {volume}
  {9}},\ \bibinfo {pages} {2104076} (\bibinfo {year} {2022})}\BibitemShut
  {NoStop}%
\bibitem [{\citenamefont {Du}\ \emph {et~al.}(2017)\citenamefont {Du},
  \citenamefont {Cai}, \citenamefont {Zidan}, \citenamefont {Ma}, \citenamefont
  {Lee},\ and\ \citenamefont {Lu}}]{du2017reservoir}%
  \BibitemOpen
  \bibfield  {author} {\bibinfo {author} {\bibfnamefont {C.}~\bibnamefont
  {Du}}, \bibinfo {author} {\bibfnamefont {F.}~\bibnamefont {Cai}}, \bibinfo
  {author} {\bibfnamefont {M.~A.}\ \bibnamefont {Zidan}}, \bibinfo {author}
  {\bibfnamefont {W.}~\bibnamefont {Ma}}, \bibinfo {author} {\bibfnamefont
  {S.~H.}\ \bibnamefont {Lee}},\ and\ \bibinfo {author} {\bibfnamefont {W.~D.}\
  \bibnamefont {Lu}},\ }\bibfield  {title} {\bibinfo {title} {Reservoir
  computing using dynamic memristors for temporal information processing},\
  }\href {https://doi.org/10.1038/s41467-017-02337-y} {\bibfield  {journal}
  {\bibinfo  {journal} {Nat. Commun.}\ }\textbf {\bibinfo {volume} {8}},\
  \bibinfo {pages} {2204} (\bibinfo {year} {2017})}\BibitemShut {NoStop}%
\bibitem [{\citenamefont {Rafayelyan}\ \emph {et~al.}(2020)\citenamefont
  {Rafayelyan}, \citenamefont {Dong}, \citenamefont {Tan}, \citenamefont
  {Krzakala},\ and\ \citenamefont {Gigan}}]{rafayelyan2020large}%
  \BibitemOpen
  \bibfield  {author} {\bibinfo {author} {\bibfnamefont {M.}~\bibnamefont
  {Rafayelyan}}, \bibinfo {author} {\bibfnamefont {J.}~\bibnamefont {Dong}},
  \bibinfo {author} {\bibfnamefont {Y.}~\bibnamefont {Tan}}, \bibinfo {author}
  {\bibfnamefont {F.}~\bibnamefont {Krzakala}},\ and\ \bibinfo {author}
  {\bibfnamefont {S.}~\bibnamefont {Gigan}},\ }\bibfield  {title} {\bibinfo
  {title} {Large-scale optical reservoir computing for spatiotemporal chaotic
  systems prediction},\ }\href {https://doi.org/10.1103/PhysRevX.10.041037}
  {\bibfield  {journal} {\bibinfo  {journal} {Phys. Rev. X}\ }\textbf {\bibinfo
  {volume} {10}},\ \bibinfo {pages} {041037} (\bibinfo {year}
  {2020})}\BibitemShut {NoStop}%
\bibitem [{\citenamefont {Jaeger}\ and\ \citenamefont
  {Haas}(2004)}]{jaeger2004harnessing}%
  \BibitemOpen
  \bibfield  {author} {\bibinfo {author} {\bibfnamefont {H.}~\bibnamefont
  {Jaeger}}\ and\ \bibinfo {author} {\bibfnamefont {H.}~\bibnamefont {Haas}},\
  }\bibfield  {title} {\bibinfo {title} {Harnessing nonlinearity: Predicting
  chaotic systems and saving energy in wireless communication},\ }\href
  {https://doi.org/10.1126/science.1091277} {\bibfield  {journal} {\bibinfo
  {journal} {Science}\ }\textbf {\bibinfo {volume} {304}},\ \bibinfo {pages}
  {78} (\bibinfo {year} {2004})}\BibitemShut {NoStop}%
\bibitem [{\citenamefont {Milano}\ \emph {et~al.}(2022)\citenamefont {Milano},
  \citenamefont {Pedretti}, \citenamefont {Montano}, \citenamefont {Ricci},
  \citenamefont {Hashemkhani}, \citenamefont {Boarino}, \citenamefont
  {Ielmini},\ and\ \citenamefont {Ricciardi}}]{milano2022materia}%
  \BibitemOpen
  \bibfield  {author} {\bibinfo {author} {\bibfnamefont {G.}~\bibnamefont
  {Milano}}, \bibinfo {author} {\bibfnamefont {G.}~\bibnamefont {Pedretti}},
  \bibinfo {author} {\bibfnamefont {K.}~\bibnamefont {Montano}}, \bibinfo
  {author} {\bibfnamefont {S.}~\bibnamefont {Ricci}}, \bibinfo {author}
  {\bibfnamefont {S.}~\bibnamefont {Hashemkhani}}, \bibinfo {author}
  {\bibfnamefont {L.}~\bibnamefont {Boarino}}, \bibinfo {author} {\bibfnamefont
  {D.}~\bibnamefont {Ielmini}},\ and\ \bibinfo {author} {\bibfnamefont
  {C.}~\bibnamefont {Ricciardi}},\ }\bibfield  {title} {\bibinfo {title} {In
  materia reservoir computing with a fully memristive architecture based on
  self-organizing nanowire networks},\ }\href
  {https://doi.org/10.1038/s41563-021-01099-9} {\bibfield  {journal} {\bibinfo
  {journal} {Nat. Mater.}\ }\textbf {\bibinfo {volume} {21}},\ \bibinfo {pages}
  {195} (\bibinfo {year} {2022})}\BibitemShut {NoStop}%
\bibitem [{\citenamefont {Pathak}\ \emph {et~al.}(2018)\citenamefont {Pathak},
  \citenamefont {Hunt}, \citenamefont {Girvan}, \citenamefont {Lu},\ and\
  \citenamefont {Ott}}]{pathak2018model}%
  \BibitemOpen
  \bibfield  {author} {\bibinfo {author} {\bibfnamefont {J.}~\bibnamefont
  {Pathak}}, \bibinfo {author} {\bibfnamefont {B.}~\bibnamefont {Hunt}},
  \bibinfo {author} {\bibfnamefont {M.}~\bibnamefont {Girvan}}, \bibinfo
  {author} {\bibfnamefont {Z.}~\bibnamefont {Lu}},\ and\ \bibinfo {author}
  {\bibfnamefont {E.}~\bibnamefont {Ott}},\ }\bibfield  {title} {\bibinfo
  {title} {Model-free prediction of large spatiotemporally chaotic systems from
  data: A reservoir computing approach},\ }\href
  {https://doi.org/10.1103/PhysRevLett.120.024102} {\bibfield  {journal}
  {\bibinfo  {journal} {Phys. Rev. Lett.}\ }\textbf {\bibinfo {volume} {120}},\
  \bibinfo {pages} {024102} (\bibinfo {year} {2018})}\BibitemShut {NoStop}%
\bibitem [{\citenamefont {Yan}(2022)}]{yan2022reservoir}%
  \BibitemOpen
  \bibfield  {author} {\bibinfo {author} {\bibfnamefont {X.}~\bibnamefont
  {Yan}},\ }\bibfield  {title} {\bibinfo {title} {Reservoir computing goes
  fully analogue},\ }\href {https://doi.org/10.1038/s41928-022-00854-3}
  {\bibfield  {journal} {\bibinfo  {journal} {Nature Electronics}\ }\textbf
  {\bibinfo {volume} {5}},\ \bibinfo {pages} {629} (\bibinfo {year}
  {2022})}\BibitemShut {NoStop}%
\bibitem [{\citenamefont {Cucchi}\ \emph {et~al.}(2021)\citenamefont {Cucchi},
  \citenamefont {Gruener}, \citenamefont {Petrauskas}, \citenamefont {Steiner},
  \citenamefont {Tseng}, \citenamefont {Fischer}, \citenamefont {Penkovsky},
  \citenamefont {Matthus}, \citenamefont {Birkholz}, \citenamefont {Kleemann},\
  and\ \citenamefont {Leo}}]{cucchi2021reservoir}%
  \BibitemOpen
  \bibfield  {author} {\bibinfo {author} {\bibfnamefont {M.}~\bibnamefont
  {Cucchi}}, \bibinfo {author} {\bibfnamefont {C.}~\bibnamefont {Gruener}},
  \bibinfo {author} {\bibfnamefont {L.}~\bibnamefont {Petrauskas}}, \bibinfo
  {author} {\bibfnamefont {P.}~\bibnamefont {Steiner}}, \bibinfo {author}
  {\bibfnamefont {H.}~\bibnamefont {Tseng}}, \bibinfo {author} {\bibfnamefont
  {A.}~\bibnamefont {Fischer}}, \bibinfo {author} {\bibfnamefont
  {B.}~\bibnamefont {Penkovsky}}, \bibinfo {author} {\bibfnamefont
  {C.}~\bibnamefont {Matthus}}, \bibinfo {author} {\bibfnamefont
  {P.}~\bibnamefont {Birkholz}}, \bibinfo {author} {\bibfnamefont
  {H.}~\bibnamefont {Kleemann}},\ and\ \bibinfo {author} {\bibfnamefont
  {K.}~\bibnamefont {Leo}},\ }\bibfield  {title} {\bibinfo {title} {Reservoir
  computing with biocompatible organic electrochemical networks for
  brain-inspired biosignal classification},\ }\href
  {https://doi.org/10.1126/sciadv.abh0693} {\bibfield  {journal} {\bibinfo
  {journal} {Sci. Adv.}\ }\textbf {\bibinfo {volume} {7}},\ \bibinfo {pages}
  {eabh0693} (\bibinfo {year} {2021})}\BibitemShut {NoStop}%
\bibitem [{\citenamefont {Pathak}\ \emph {et~al.}(2017)\citenamefont {Pathak},
  \citenamefont {Lu}, \citenamefont {Hunt}, \citenamefont {Girvan},\ and\
  \citenamefont {Ott}}]{pathak2017using}%
  \BibitemOpen
  \bibfield  {author} {\bibinfo {author} {\bibfnamefont {J.}~\bibnamefont
  {Pathak}}, \bibinfo {author} {\bibfnamefont {Z.}~\bibnamefont {Lu}}, \bibinfo
  {author} {\bibfnamefont {B.~R.}\ \bibnamefont {Hunt}}, \bibinfo {author}
  {\bibfnamefont {M.}~\bibnamefont {Girvan}},\ and\ \bibinfo {author}
  {\bibfnamefont {E.}~\bibnamefont {Ott}},\ }\bibfield  {title} {\bibinfo
  {title} {Using machine learning to replicate chaotic attractors and calculate
  lyapunov exponents from data},\ }\href {https://doi.org/10.1063/1.5010300}
  {\bibfield  {journal} {\bibinfo  {journal} {Chaos}\ }\textbf {\bibinfo
  {volume} {27}},\ \bibinfo {pages} {121102} (\bibinfo {year}
  {2017})}\BibitemShut {NoStop}%
\bibitem [{\citenamefont {Panahi}\ and\ \citenamefont
  {Lai}(2024)}]{panahi2024adaptable}%
  \BibitemOpen
  \bibfield  {author} {\bibinfo {author} {\bibfnamefont {S.}~\bibnamefont
  {Panahi}}\ and\ \bibinfo {author} {\bibfnamefont {Y.-C.}\ \bibnamefont
  {Lai}},\ }\bibfield  {title} {\bibinfo {title} {Adaptable reservoir
  computing: A paradigm for model-free data-driven prediction of critical
  transitions in nonlinear dynamical systems},\ }\href
  {https://doi.org/10.1063/5.0200898} {\bibfield  {journal} {\bibinfo
  {journal} {Chaos}\ }\textbf {\bibinfo {volume} {34}},\ \bibinfo {pages}
  {051501} (\bibinfo {year} {2024})}\BibitemShut {NoStop}%
\bibitem [{\citenamefont {Fang}\ and\ \citenamefont
  {Mengaldo}(2025)}]{fang2025dynamical}%
  \BibitemOpen
  \bibfield  {author} {\bibinfo {author} {\bibfnamefont {Z.}~\bibnamefont
  {Fang}}\ and\ \bibinfo {author} {\bibfnamefont {G.}~\bibnamefont
  {Mengaldo}},\ }\href {https://arxiv.org/abs/2504.11074} {\bibinfo {title}
  {Dynamical errors in machine learning forecasts}} (\bibinfo {year} {2025}),\
  \Eprint {https://arxiv.org/abs/2504.11074} {arXiv:2504.11074 [cs.LG]}
  \BibitemShut {NoStop}%
\bibitem [{\citenamefont {Xiao}\ \emph {et~al.}(2021)\citenamefont {Xiao},
  \citenamefont {Kong}, \citenamefont {Sun},\ and\ \citenamefont
  {Lai}}]{xiao2021predicting}%
  \BibitemOpen
  \bibfield  {author} {\bibinfo {author} {\bibfnamefont {R.}~\bibnamefont
  {Xiao}}, \bibinfo {author} {\bibfnamefont {L.-W.}\ \bibnamefont {Kong}},
  \bibinfo {author} {\bibfnamefont {Z.-K.}\ \bibnamefont {Sun}},\ and\ \bibinfo
  {author} {\bibfnamefont {Y.-C.}\ \bibnamefont {Lai}},\ }\bibfield  {title}
  {\bibinfo {title} {Predicting amplitude death with machine learning},\ }\href
  {https://doi.org/10.1103/PhysRevE.104.014205} {\bibfield  {journal} {\bibinfo
   {journal} {Phys. Rev. E}\ }\textbf {\bibinfo {volume} {104}},\ \bibinfo
  {pages} {014205} (\bibinfo {year} {2021})}\BibitemShut {NoStop}%
\bibitem [{\citenamefont {Gauthier}\ \emph {et~al.}(2021)\citenamefont
  {Gauthier}, \citenamefont {Bollt}, \citenamefont {Griffith},\ and\
  \citenamefont {Barbosa}}]{gauthier2021next}%
  \BibitemOpen
  \bibfield  {author} {\bibinfo {author} {\bibfnamefont {D.~J.}\ \bibnamefont
  {Gauthier}}, \bibinfo {author} {\bibfnamefont {E.}~\bibnamefont {Bollt}},
  \bibinfo {author} {\bibfnamefont {A.}~\bibnamefont {Griffith}},\ and\
  \bibinfo {author} {\bibfnamefont {W.~A.~S.}\ \bibnamefont {Barbosa}},\
  }\bibfield  {title} {\bibinfo {title} {Next generation reservoir computing},\
  }\href {https://doi.org/10.1038/s41467-021-25801-2} {\bibfield  {journal}
  {\bibinfo  {journal} {Nat. Commun.}\ }\textbf {\bibinfo {volume} {12}},\
  \bibinfo {pages} {5564} (\bibinfo {year} {2021})}\BibitemShut {NoStop}%
\bibitem [{\citenamefont {Wikner}\ \emph {et~al.}(2021)\citenamefont {Wikner},
  \citenamefont {Pathak}, \citenamefont {Hunt}, \citenamefont {Szunyogh},
  \citenamefont {Girvan},\ and\ \citenamefont {Ott}}]{2021-chaos}%
  \BibitemOpen
  \bibfield  {author} {\bibinfo {author} {\bibfnamefont {A.}~\bibnamefont
  {Wikner}}, \bibinfo {author} {\bibfnamefont {J.}~\bibnamefont {Pathak}},
  \bibinfo {author} {\bibfnamefont {B.~R.}\ \bibnamefont {Hunt}}, \bibinfo
  {author} {\bibfnamefont {I.}~\bibnamefont {Szunyogh}}, \bibinfo {author}
  {\bibfnamefont {M.}~\bibnamefont {Girvan}},\ and\ \bibinfo {author}
  {\bibfnamefont {E.}~\bibnamefont {Ott}},\ }\bibfield  {title} {\bibinfo
  {title} {Using data assimilation to train a hybrid forecast system that
  combines machine-learning and knowledge-based components},\ }\href
  {https://doi.org/10.1063/5.0048050} {\bibfield  {journal} {\bibinfo
  {journal} {Chaos}\ }\textbf {\bibinfo {volume} {31}},\ \bibinfo {pages}
  {053114} (\bibinfo {year} {2021})}\BibitemShut {NoStop}%
\bibitem [{\citenamefont {Chepuri}\ \emph {et~al.}(2024)\citenamefont
  {Chepuri}, \citenamefont {Amzalag}, \citenamefont {Antonsen},\ and\
  \citenamefont {Girvan}}]{chepuri2024hybridizing}%
  \BibitemOpen
  \bibfield  {author} {\bibinfo {author} {\bibfnamefont {R.}~\bibnamefont
  {Chepuri}}, \bibinfo {author} {\bibfnamefont {D.}~\bibnamefont {Amzalag}},
  \bibinfo {author} {\bibfnamefont {T.~M.}\ \bibnamefont {Antonsen}},\ and\
  \bibinfo {author} {\bibfnamefont {M.}~\bibnamefont {Girvan}},\ }\bibfield
  {title} {\bibinfo {title} {Hybridizing traditional and next-generation
  reservoir computing to accurately and efficiently forecast dynamical
  systems},\ }\href {https://doi.org/10.1063/5.0206232} {\bibfield  {journal}
  {\bibinfo  {journal} {Chaos}\ }\textbf {\bibinfo {volume} {34}},\ \bibinfo
  {pages} {063114} (\bibinfo {year} {2024})}\BibitemShut {NoStop}%
\bibitem [{\citenamefont {Wang}\ \emph
  {et~al.}(2024{\natexlab{a}})\citenamefont {Wang}, \citenamefont {Sun},\ and\
  \citenamefont {Zhang}}]{wang2024enhanced}%
  \BibitemOpen
  \bibfield  {author} {\bibinfo {author} {\bibfnamefont {L.}~\bibnamefont
  {Wang}}, \bibinfo {author} {\bibfnamefont {Y.}~\bibnamefont {Sun}},\ and\
  \bibinfo {author} {\bibfnamefont {X.}~\bibnamefont {Zhang}},\ }\bibfield
  {title} {\bibinfo {title} {Enhanced predictive capability for chaotic
  dynamics by modified quantum reservoir computing},\ }\href
  {https://doi.org/10.1103/PhysRevResearch.6.043183} {\bibfield  {journal}
  {\bibinfo  {journal} {Phys. Rev. Res.}\ }\textbf {\bibinfo {volume} {6}},\
  \bibinfo {pages} {043183} (\bibinfo {year} {2024}{\natexlab{a}})}\BibitemShut
  {NoStop}%
\bibitem [{\citenamefont {Ramadevi}\ and\ \citenamefont
  {Bingi}(2022)}]{ramadevi2022chaotic}%
  \BibitemOpen
  \bibfield  {author} {\bibinfo {author} {\bibfnamefont {B.}~\bibnamefont
  {Ramadevi}}\ and\ \bibinfo {author} {\bibfnamefont {K.}~\bibnamefont
  {Bingi}},\ }\bibfield  {title} {\bibinfo {title} {Chaotic time series
  forecasting approaches using machine learning techniques: A review},\ }\href
  {https://doi.org/10.3390/sym14050955} {\bibfield  {journal} {\bibinfo
  {journal} {Symmetry}\ }\textbf {\bibinfo {volume} {14}},\ \bibinfo {pages}
  {955} (\bibinfo {year} {2022})}\BibitemShut {NoStop}%
\bibitem [{\citenamefont {Hochreiter}\ and\ \citenamefont
  {Schmidhuber}(1997)}]{hochreiter1997long}%
  \BibitemOpen
  \bibfield  {author} {\bibinfo {author} {\bibfnamefont {S.}~\bibnamefont
  {Hochreiter}}\ and\ \bibinfo {author} {\bibfnamefont {J.}~\bibnamefont
  {Schmidhuber}},\ }\bibfield  {title} {\bibinfo {title} {Long short-term
  memory},\ }\href {https://doi.org/10.1162/neco.1997.9.8.1735} {\bibfield
  {journal} {\bibinfo  {journal} {Neural Comput.}\ }\textbf {\bibinfo {volume}
  {9}},\ \bibinfo {pages} {1735} (\bibinfo {year} {1997})}\BibitemShut
  {NoStop}%
\bibitem [{\citenamefont {Landi}\ \emph {et~al.}(2021)\citenamefont {Landi},
  \citenamefont {Baraldi}, \citenamefont {Cornia},\ and\ \citenamefont
  {Cucchiara}}]{landi2021working}%
  \BibitemOpen
  \bibfield  {author} {\bibinfo {author} {\bibfnamefont {F.}~\bibnamefont
  {Landi}}, \bibinfo {author} {\bibfnamefont {L.}~\bibnamefont {Baraldi}},
  \bibinfo {author} {\bibfnamefont {M.}~\bibnamefont {Cornia}},\ and\ \bibinfo
  {author} {\bibfnamefont {R.}~\bibnamefont {Cucchiara}},\ }\bibfield  {title}
  {\bibinfo {title} {Working memory connections for lstm},\ }\href
  {https://doi.org/https://doi.org/10.1016/j.neunet.2021.08.030} {\bibfield
  {journal} {\bibinfo  {journal} {Neural Networks}\ }\textbf {\bibinfo {volume}
  {144}},\ \bibinfo {pages} {334} (\bibinfo {year} {2021})}\BibitemShut
  {NoStop}%
\bibitem [{\citenamefont {Shahi}\ \emph {et~al.}(2022)\citenamefont {Shahi},
  \citenamefont {Fenton},\ and\ \citenamefont {Cherry}}]{shahi2022prediction}%
  \BibitemOpen
  \bibfield  {author} {\bibinfo {author} {\bibfnamefont {S.}~\bibnamefont
  {Shahi}}, \bibinfo {author} {\bibfnamefont {F.~H.}\ \bibnamefont {Fenton}},\
  and\ \bibinfo {author} {\bibfnamefont {E.~M.}\ \bibnamefont {Cherry}},\
  }\bibfield  {title} {\bibinfo {title} {Prediction of chaotic time series
  using recurrent neural networks and reservoir computing techniques: A
  comparative study},\ }\href
  {https://doi.org/https://doi.org/10.1016/j.mlwa.2022.100300} {\bibfield
  {journal} {\bibinfo  {journal} {Machine Learning with Applications}\ }\textbf
  {\bibinfo {volume} {8}},\ \bibinfo {pages} {100300} (\bibinfo {year}
  {2022})}\BibitemShut {NoStop}%
\bibitem [{\citenamefont {Wang}\ \emph {et~al.}(2023)\citenamefont {Wang},
  \citenamefont {Shao},\ and\ \citenamefont {Jumahong}}]{Wang2023_FuzzyLSTM}%
  \BibitemOpen
  \bibfield  {author} {\bibinfo {author} {\bibfnamefont {W.}~\bibnamefont
  {Wang}}, \bibinfo {author} {\bibfnamefont {J.}~\bibnamefont {Shao}},\ and\
  \bibinfo {author} {\bibfnamefont {H.}~\bibnamefont {Jumahong}},\ }\bibfield
  {title} {\bibinfo {title} {Fuzzy inference-based lstm for long-term
  time-series prediction},\ }\href {https://doi.org/10.1038/s41598-023-47812-3}
  {\bibfield  {journal} {\bibinfo  {journal} {Sci. Rep.}\ }\textbf {\bibinfo
  {volume} {13}},\ \bibinfo {pages} {20359} (\bibinfo {year}
  {2023})}\BibitemShut {NoStop}%
\bibitem [{\citenamefont {Bai}\ \emph {et~al.}(2018)\citenamefont {Bai},
  \citenamefont {Kolter},\ and\ \citenamefont {Koltun}}]{bai2018empirical}%
  \BibitemOpen
  \bibfield  {author} {\bibinfo {author} {\bibfnamefont {S.}~\bibnamefont
  {Bai}}, \bibinfo {author} {\bibfnamefont {J.~Z.}\ \bibnamefont {Kolter}},\
  and\ \bibinfo {author} {\bibfnamefont {V.}~\bibnamefont {Koltun}},\ }\href
  {https://arxiv.org/abs/1803.01271} {\bibinfo {title} {An empirical evaluation
  of generic convolutional and recurrent networks for sequence modeling}}
  (\bibinfo {year} {2018}),\ \Eprint {https://arxiv.org/abs/1803.01271}
  {arXiv:1803.01271 [cs.LG]} \BibitemShut {NoStop}%
\bibitem [{\citenamefont {Tiezzi}\ \emph {et~al.}(2025)\citenamefont {Tiezzi},
  \citenamefont {Casoni}, \citenamefont {Betti}, \citenamefont {Guidi},
  \citenamefont {Gori},\ and\ \citenamefont {Melacci}}]{tiezzi2025back}%
  \BibitemOpen
  \bibfield  {author} {\bibinfo {author} {\bibfnamefont {M.}~\bibnamefont
  {Tiezzi}}, \bibinfo {author} {\bibfnamefont {M.}~\bibnamefont {Casoni}},
  \bibinfo {author} {\bibfnamefont {A.}~\bibnamefont {Betti}}, \bibinfo
  {author} {\bibfnamefont {T.}~\bibnamefont {Guidi}}, \bibinfo {author}
  {\bibfnamefont {M.}~\bibnamefont {Gori}},\ and\ \bibinfo {author}
  {\bibfnamefont {S.}~\bibnamefont {Melacci}},\ }\bibfield  {title} {\bibinfo
  {title} {Back to recurrent processing at the crossroad of transformers and
  state-space models},\ }\href {https://doi.org/10.1038/s42256-025-01034-6}
  {\bibfield  {journal} {\bibinfo  {journal} {Nature Machine Intelligence}\
  }\textbf {\bibinfo {volume} {7}},\ \bibinfo {pages} {678} (\bibinfo {year}
  {2025})}\BibitemShut {NoStop}%
\bibitem [{\citenamefont {Vaswani}\ \emph {et~al.}(2017)\citenamefont
  {Vaswani}, \citenamefont {Shazeer}, \citenamefont {Parmar}, \citenamefont
  {Uszkoreit}, \citenamefont {Jones}, \citenamefont {Gomez}, \citenamefont
  {Kaiser},\ and\ \citenamefont {Polosukhin}}]{ashish2017attention}%
  \BibitemOpen
  \bibfield  {author} {\bibinfo {author} {\bibfnamefont {A.}~\bibnamefont
  {Vaswani}}, \bibinfo {author} {\bibfnamefont {N.}~\bibnamefont {Shazeer}},
  \bibinfo {author} {\bibfnamefont {N.}~\bibnamefont {Parmar}}, \bibinfo
  {author} {\bibfnamefont {J.}~\bibnamefont {Uszkoreit}}, \bibinfo {author}
  {\bibfnamefont {L.}~\bibnamefont {Jones}}, \bibinfo {author} {\bibfnamefont
  {A.~N.}\ \bibnamefont {Gomez}}, \bibinfo {author} {\bibfnamefont {L.~u.}\
  \bibnamefont {Kaiser}},\ and\ \bibinfo {author} {\bibfnamefont
  {I.}~\bibnamefont {Polosukhin}},\ }\bibfield  {title} {\bibinfo {title}
  {Attention is all you need},\ }in\ \href
  {https://proceedings.neurips.cc/paper_files/paper/2017/file/3f5ee243547dee91fbd053c1c4a845aa-Paper.pdf}
  {\emph {\bibinfo {booktitle} {Advances in Neural Information Processing
  Systems}}},\ Vol.~\bibinfo {volume} {30},\ \bibinfo {editor} {edited by\
  \bibinfo {editor} {\bibfnamefont {I.}~\bibnamefont {Guyon}}, \bibinfo
  {editor} {\bibfnamefont {U.~V.}\ \bibnamefont {Luxburg}}, \bibinfo {editor}
  {\bibfnamefont {S.}~\bibnamefont {Bengio}}, \bibinfo {editor} {\bibfnamefont
  {H.}~\bibnamefont {Wallach}}, \bibinfo {editor} {\bibfnamefont
  {R.}~\bibnamefont {Fergus}}, \bibinfo {editor} {\bibfnamefont
  {S.}~\bibnamefont {Vishwanathan}},\ and\ \bibinfo {editor} {\bibfnamefont
  {R.}~\bibnamefont {Garnett}}}\ (\bibinfo  {publisher} {Curran Associates,
  Inc.},\ \bibinfo {year} {2017})\BibitemShut {NoStop}%
\bibitem [{\citenamefont {Zhou}\ \emph {et~al.}(2021)\citenamefont {Zhou},
  \citenamefont {Zhang}, \citenamefont {Peng}, \citenamefont {Zhang},
  \citenamefont {Li}, \citenamefont {Xiong},\ and\ \citenamefont
  {Zhang}}]{zhou2021informer}%
  \BibitemOpen
  \bibfield  {author} {\bibinfo {author} {\bibfnamefont {H.}~\bibnamefont
  {Zhou}}, \bibinfo {author} {\bibfnamefont {S.}~\bibnamefont {Zhang}},
  \bibinfo {author} {\bibfnamefont {J.}~\bibnamefont {Peng}}, \bibinfo {author}
  {\bibfnamefont {S.}~\bibnamefont {Zhang}}, \bibinfo {author} {\bibfnamefont
  {J.}~\bibnamefont {Li}}, \bibinfo {author} {\bibfnamefont {H.}~\bibnamefont
  {Xiong}},\ and\ \bibinfo {author} {\bibfnamefont {W.}~\bibnamefont {Zhang}},\
  }\bibfield  {title} {\bibinfo {title} {Informer: Beyond efficient transformer
  for long sequence time-series forecasting},\ }\href
  {https://doi.org/10.1609/aaai.v35i12.17325} {\bibfield  {journal} {\bibinfo
  {journal} {Proceedings of the AAAI Conference on Artificial Intelligence}\
  }\textbf {\bibinfo {volume} {35}},\ \bibinfo {pages} {11106} (\bibinfo {year}
  {2021})}\BibitemShut {NoStop}%
\bibitem [{\citenamefont {Wu}\ \emph {et~al.}(2021)\citenamefont {Wu},
  \citenamefont {Xu}, \citenamefont {Wang},\ and\ \citenamefont
  {Long}}]{wu2021autoformer}%
  \BibitemOpen
  \bibfield  {author} {\bibinfo {author} {\bibfnamefont {H.}~\bibnamefont
  {Wu}}, \bibinfo {author} {\bibfnamefont {J.}~\bibnamefont {Xu}}, \bibinfo
  {author} {\bibfnamefont {J.}~\bibnamefont {Wang}},\ and\ \bibinfo {author}
  {\bibfnamefont {M.}~\bibnamefont {Long}},\ }\bibfield  {title} {\bibinfo
  {title} {Autoformer: Decomposition transformers with auto-correlation for
  long-term series forecasting},\ }in\ \href
  {https://proceedings.neurips.cc/paper_files/paper/2021/file/bcc0d400288793e8bdcd7c19a8ac0c2b-Paper.pdf}
  {\emph {\bibinfo {booktitle} {Advances in Neural Information Processing
  Systems}}},\ Vol.~\bibinfo {volume} {34},\ \bibinfo {editor} {edited by\
  \bibinfo {editor} {\bibfnamefont {M.}~\bibnamefont {Ranzato}}, \bibinfo
  {editor} {\bibfnamefont {A.}~\bibnamefont {Beygelzimer}}, \bibinfo {editor}
  {\bibfnamefont {Y.}~\bibnamefont {Dauphin}}, \bibinfo {editor} {\bibfnamefont
  {P.}~\bibnamefont {Liang}},\ and\ \bibinfo {editor} {\bibfnamefont {J.~W.}\
  \bibnamefont {Vaughan}}}\ (\bibinfo  {publisher} {Curran Associates, Inc.},\
  \bibinfo {year} {2021})\ pp.\ \bibinfo {pages} {22419--22430}\BibitemShut
  {NoStop}%
\bibitem [{\citenamefont {Zeng}\ \emph {et~al.}(2022)\citenamefont {Zeng},
  \citenamefont {Chen}, \citenamefont {Zhang},\ and\ \citenamefont
  {Xu}}]{zeng2023transformers}%
  \BibitemOpen
  \bibfield  {author} {\bibinfo {author} {\bibfnamefont {A.}~\bibnamefont
  {Zeng}}, \bibinfo {author} {\bibfnamefont {M.}~\bibnamefont {Chen}}, \bibinfo
  {author} {\bibfnamefont {L.}~\bibnamefont {Zhang}},\ and\ \bibinfo {author}
  {\bibfnamefont {Q.}~\bibnamefont {Xu}},\ }\bibfield  {title} {\bibinfo
  {title} {Are transformers effective for time series forecasting?}\ }(\bibinfo
  {year} {2022})\BibitemShut {NoStop}%
\bibitem [{\citenamefont {Wu}\ \emph {et~al.}(2023)\citenamefont {Wu},
  \citenamefont {Hu}, \citenamefont {Liu}, \citenamefont {Zhou}, \citenamefont
  {Wang},\ and\ \citenamefont {Long}}]{wu2022timesnet}%
  \BibitemOpen
  \bibfield  {author} {\bibinfo {author} {\bibfnamefont {H.}~\bibnamefont
  {Wu}}, \bibinfo {author} {\bibfnamefont {T.}~\bibnamefont {Hu}}, \bibinfo
  {author} {\bibfnamefont {Y.}~\bibnamefont {Liu}}, \bibinfo {author}
  {\bibfnamefont {H.}~\bibnamefont {Zhou}}, \bibinfo {author} {\bibfnamefont
  {J.}~\bibnamefont {Wang}},\ and\ \bibinfo {author} {\bibfnamefont
  {M.}~\bibnamefont {Long}},\ }\href {https://arxiv.org/abs/2210.02186}
  {\bibinfo {title} {Timesnet: Temporal 2d-variation modeling for general time
  series analysis}} (\bibinfo {year} {2023}),\ \Eprint
  {https://arxiv.org/abs/2210.02186} {arXiv:2210.02186 [cs.LG]} \BibitemShut
  {NoStop}%
\bibitem [{\citenamefont {Lim}\ \emph {et~al.}(2021)\citenamefont {Lim},
  \citenamefont {Arık}, \citenamefont {Loeff},\ and\ \citenamefont
  {Pfister}}]{lim2021temporal}%
  \BibitemOpen
  \bibfield  {author} {\bibinfo {author} {\bibfnamefont {B.}~\bibnamefont
  {Lim}}, \bibinfo {author} {\bibfnamefont {S.~.}\ \bibnamefont {Arık}},
  \bibinfo {author} {\bibfnamefont {N.}~\bibnamefont {Loeff}},\ and\ \bibinfo
  {author} {\bibfnamefont {T.}~\bibnamefont {Pfister}},\ }\bibfield  {title}
  {\bibinfo {title} {Temporal fusion transformers for interpretable
  multi-horizon time series forecasting},\ }\href
  {https://doi.org/https://doi.org/10.1016/j.ijforecast.2021.03.012} {\bibfield
   {journal} {\bibinfo  {journal} {Int. J. Forecasting}\ }\textbf {\bibinfo
  {volume} {37}},\ \bibinfo {pages} {1748} (\bibinfo {year}
  {2021})}\BibitemShut {NoStop}%
\bibitem [{\citenamefont {Cao}\ \emph {et~al.}(2025)\citenamefont {Cao},
  \citenamefont {Meng}, \citenamefont {Li}, \citenamefont {Wu},\ and\
  \citenamefont {Fan}}]{cao2024remaining}%
  \BibitemOpen
  \bibfield  {author} {\bibinfo {author} {\bibfnamefont {W.}~\bibnamefont
  {Cao}}, \bibinfo {author} {\bibfnamefont {Z.}~\bibnamefont {Meng}}, \bibinfo
  {author} {\bibfnamefont {J.}~\bibnamefont {Li}}, \bibinfo {author}
  {\bibfnamefont {J.}~\bibnamefont {Wu}},\ and\ \bibinfo {author}
  {\bibfnamefont {F.}~\bibnamefont {Fan}},\ }\bibfield  {title} {\bibinfo
  {title} {A remaining useful life prediction method for rolling bearing based
  on tcn-transformer},\ }\href {https://doi.org/10.1109/TIM.2024.3502878}
  {\bibfield  {journal} {\bibinfo  {journal} {IEEE Trans. Instrum. Meas.}\
  }\textbf {\bibinfo {volume} {74}},\ \bibinfo {pages} {1} (\bibinfo {year}
  {2025})}\BibitemShut {NoStop}%
\bibitem [{\citenamefont {Kabir}\ \emph {et~al.}(2025)\citenamefont {Kabir},
  \citenamefont {Bhadra}, \citenamefont {Ridoy},\ and\ \citenamefont
  {Milanova}}]{sci7010007}%
  \BibitemOpen
  \bibfield  {author} {\bibinfo {author} {\bibfnamefont {M.~R.}\ \bibnamefont
  {Kabir}}, \bibinfo {author} {\bibfnamefont {D.}~\bibnamefont {Bhadra}},
  \bibinfo {author} {\bibfnamefont {M.}~\bibnamefont {Ridoy}},\ and\ \bibinfo
  {author} {\bibfnamefont {M.}~\bibnamefont {Milanova}},\ }\bibfield  {title}
  {\bibinfo {title} {Lstm–transformer-based robust hybrid deep learning model
  for financial time series forecasting},\ }\bibfield  {journal} {\bibinfo
  {journal} {Sci}\ }\textbf {\bibinfo {volume} {7}},\ \href
  {https://doi.org/10.3390/sci7010007} {10.3390/sci7010007} (\bibinfo {year}
  {2025})\BibitemShut {NoStop}%
\bibitem [{\citenamefont {Guo}\ and\ \citenamefont {Li}(2025)}]{2025Guo}%
  \BibitemOpen
  \bibfield  {author} {\bibinfo {author} {\bibfnamefont {W.}~\bibnamefont
  {Guo}}\ and\ \bibinfo {author} {\bibfnamefont {M.}~\bibnamefont {Li}},\
  }\bibfield  {title} {\bibinfo {title} {A hybrid model based on
  transformer-lstm for battery health prediction},\ }in\ \href
  {https://doi.org/10.1109/ICEPET65469.2025.11047391} {\emph {\bibinfo
  {booktitle} {2025 4th International Conference on Energy, Power and
  Electrical Technology (ICEPET)}}}\ (\bibinfo {year} {2025})\ pp.\ \bibinfo
  {pages} {1001--1004}\BibitemShut {NoStop}%
\bibitem [{\citenamefont {Viehweg}\ \emph {et~al.}(2025)\citenamefont
  {Viehweg}, \citenamefont {Poll},\ and\ \citenamefont
  {M{\"a}der}}]{viehweg2025deterministic}%
  \BibitemOpen
  \bibfield  {author} {\bibinfo {author} {\bibfnamefont {J.}~\bibnamefont
  {Viehweg}}, \bibinfo {author} {\bibfnamefont {C.}~\bibnamefont {Poll}},\ and\
  \bibinfo {author} {\bibfnamefont {P.}~\bibnamefont {M{\"a}der}},\ }\bibfield
  {title} {\bibinfo {title} {Deterministic reservoir computing for chaotic time
  series prediction},\ }\href {https://doi.org/10.1038/s41598-025-98172-z}
  {\bibfield  {journal} {\bibinfo  {journal} {Sci. Rep.}\ }\textbf {\bibinfo
  {volume} {15}},\ \bibinfo {pages} {17695} (\bibinfo {year}
  {2025})}\BibitemShut {NoStop}%
\bibitem [{\citenamefont {Hu}\ \emph {et~al.}(2024)\citenamefont {Hu},
  \citenamefont {Hu}, \citenamefont {Chen}, \citenamefont {Jin}, \citenamefont
  {Pan}, \citenamefont {Wen},\ and\ \citenamefont {Liang}}]{hu2024attractor}%
  \BibitemOpen
  \bibfield  {author} {\bibinfo {author} {\bibfnamefont {J.}~\bibnamefont
  {Hu}}, \bibinfo {author} {\bibfnamefont {Y.}~\bibnamefont {Hu}}, \bibinfo
  {author} {\bibfnamefont {W.}~\bibnamefont {Chen}}, \bibinfo {author}
  {\bibfnamefont {M.}~\bibnamefont {Jin}}, \bibinfo {author} {\bibfnamefont
  {S.}~\bibnamefont {Pan}}, \bibinfo {author} {\bibfnamefont {Q.}~\bibnamefont
  {Wen}},\ and\ \bibinfo {author} {\bibfnamefont {Y.}~\bibnamefont {Liang}},\
  }\bibfield  {title} {\bibinfo {title} {Attractor memory for long-term time
  series forecasting: A chaos perspective},\ }in\ \href
  {https://proceedings.neurips.cc/paper_files/paper/2024/file/24ef004f733548db6b3197d9f68dcb85-Paper-Conference.pdf}
  {\emph {\bibinfo {booktitle} {Advances in Neural Information Processing
  Systems}}},\ Vol.~\bibinfo {volume} {37},\ \bibinfo {editor} {edited by\
  \bibinfo {editor} {\bibfnamefont {A.}~\bibnamefont {Globerson}}, \bibinfo
  {editor} {\bibfnamefont {L.}~\bibnamefont {Mackey}}, \bibinfo {editor}
  {\bibfnamefont {D.}~\bibnamefont {Belgrave}}, \bibinfo {editor}
  {\bibfnamefont {A.}~\bibnamefont {Fan}}, \bibinfo {editor} {\bibfnamefont
  {U.}~\bibnamefont {Paquet}}, \bibinfo {editor} {\bibfnamefont
  {J.}~\bibnamefont {Tomczak}},\ and\ \bibinfo {editor} {\bibfnamefont
  {C.}~\bibnamefont {Zhang}}}\ (\bibinfo  {publisher} {Curran Associates,
  Inc.},\ \bibinfo {year} {2024})\ pp.\ \bibinfo {pages}
  {20786--20818}\BibitemShut {NoStop}%
\bibitem [{\citenamefont {Cai}\ \emph {et~al.}(2024)\citenamefont {Cai},
  \citenamefont {Mu}, \citenamefont {Huang}, \citenamefont {Zhou},\ and\
  \citenamefont {Li}}]{cai2024reinforced}%
  \BibitemOpen
  \bibfield  {author} {\bibinfo {author} {\bibfnamefont {D.}~\bibnamefont
  {Cai}}, \bibinfo {author} {\bibfnamefont {P.}~\bibnamefont {Mu}}, \bibinfo
  {author} {\bibfnamefont {Y.}~\bibnamefont {Huang}}, \bibinfo {author}
  {\bibfnamefont {P.}~\bibnamefont {Zhou}},\ and\ \bibinfo {author}
  {\bibfnamefont {N.}~\bibnamefont {Li}},\ }\bibfield  {title} {\bibinfo
  {title} {A reinforced reservoir computer aided by an external asymmetric
  dual-path-filtering cavity laser},\ }\href
  {https://doi.org/https://doi.org/10.1016/j.chaos.2024.115652} {\bibfield
  {journal} {\bibinfo  {journal} {Chaos, Solitons $\&$ Fractals}\ }\textbf
  {\bibinfo {volume} {189}},\ \bibinfo {pages} {115652} (\bibinfo {year}
  {2024})}\BibitemShut {NoStop}%
\bibitem [{\citenamefont {Wang}\ \emph
  {et~al.}(2024{\natexlab{b}})\citenamefont {Wang}, \citenamefont {Jiang},
  \citenamefont {Yan}, \citenamefont {He}, \citenamefont {Feng}, \citenamefont
  {Pan},\ and\ \citenamefont {Luo}}]{wang2024chaotic}%
  \BibitemOpen
  \bibfield  {author} {\bibinfo {author} {\bibfnamefont {Q.}~\bibnamefont
  {Wang}}, \bibinfo {author} {\bibfnamefont {L.}~\bibnamefont {Jiang}},
  \bibinfo {author} {\bibfnamefont {L.}~\bibnamefont {Yan}}, \bibinfo {author}
  {\bibfnamefont {X.}~\bibnamefont {He}}, \bibinfo {author} {\bibfnamefont
  {J.}~\bibnamefont {Feng}}, \bibinfo {author} {\bibfnamefont {W.}~\bibnamefont
  {Pan}},\ and\ \bibinfo {author} {\bibfnamefont {B.}~\bibnamefont {Luo}},\
  }\bibfield  {title} {\bibinfo {title} {Chaotic time series prediction based
  on physics-informed neural operator},\ }\href
  {https://doi.org/https://doi.org/10.1016/j.chaos.2024.115326} {\bibfield
  {journal} {\bibinfo  {journal} {Chaos, Solitons $\&$ Fractals}\ }\textbf
  {\bibinfo {volume} {186}},\ \bibinfo {pages} {115326} (\bibinfo {year}
  {2024}{\natexlab{b}})}\BibitemShut {NoStop}%
\bibitem [{\citenamefont {Yuan}\ \emph {et~al.}(2024)\citenamefont {Yuan},
  \citenamefont {Jiang}, \citenamefont {Yan}, \citenamefont {Li}, \citenamefont
  {Zhang}, \citenamefont {Yi}, \citenamefont {Pan},\ and\ \citenamefont
  {Luo}}]{yuan2024optoelectronic}%
  \BibitemOpen
  \bibfield  {author} {\bibinfo {author} {\bibfnamefont {X.}~\bibnamefont
  {Yuan}}, \bibinfo {author} {\bibfnamefont {L.}~\bibnamefont {Jiang}},
  \bibinfo {author} {\bibfnamefont {L.}~\bibnamefont {Yan}}, \bibinfo {author}
  {\bibfnamefont {S.}~\bibnamefont {Li}}, \bibinfo {author} {\bibfnamefont
  {L.}~\bibnamefont {Zhang}}, \bibinfo {author} {\bibfnamefont
  {A.}~\bibnamefont {Yi}}, \bibinfo {author} {\bibfnamefont {W.}~\bibnamefont
  {Pan}},\ and\ \bibinfo {author} {\bibfnamefont {B.}~\bibnamefont {Luo}},\
  }\bibfield  {title} {\bibinfo {title} {The optoelectronic reservoir computing
  system based on parallel multi-time-delay feedback loops for time-series
  prediction and optical performance monitoring},\ }\href
  {https://doi.org/https://doi.org/10.1016/j.chaos.2024.115306} {\bibfield
  {journal} {\bibinfo  {journal} {Chaos, Solitons $\&$ Fractals}\ }\textbf
  {\bibinfo {volume} {186}},\ \bibinfo {pages} {115306} (\bibinfo {year}
  {2024})}\BibitemShut {NoStop}%
\bibitem [{\citenamefont {Lu}\ \emph {et~al.}(2017)\citenamefont {Lu},
  \citenamefont {Pathak}, \citenamefont {Hunt}, \citenamefont {Girvan},
  \citenamefont {Brockett},\ and\ \citenamefont {Ott}}]{lu2017reservoir}%
  \BibitemOpen
  \bibfield  {author} {\bibinfo {author} {\bibfnamefont {Z.}~\bibnamefont
  {Lu}}, \bibinfo {author} {\bibfnamefont {J.}~\bibnamefont {Pathak}}, \bibinfo
  {author} {\bibfnamefont {B.}~\bibnamefont {Hunt}}, \bibinfo {author}
  {\bibfnamefont {M.}~\bibnamefont {Girvan}}, \bibinfo {author} {\bibfnamefont
  {R.}~\bibnamefont {Brockett}},\ and\ \bibinfo {author} {\bibfnamefont
  {E.}~\bibnamefont {Ott}},\ }\bibfield  {title} {\bibinfo {title} {Reservoir
  observers: Model-free inference of unmeasured variables in chaotic systems},\
  }\href {https://doi.org/10.1063/1.4979665} {\bibfield  {journal} {\bibinfo
  {journal} {Chaos}\ }\textbf {\bibinfo {volume} {27}},\ \bibinfo {pages}
  {041102} (\bibinfo {year} {2017})}\BibitemShut {NoStop}%
\bibitem [{\citenamefont {Feng}\ \emph {et~al.}(2022)\citenamefont {Feng},
  \citenamefont {Wang}, \citenamefont {Wang}, \citenamefont {Zhang},\ and\
  \citenamefont {Zhao}}]{feng2022less}%
  \BibitemOpen
  \bibfield  {author} {\bibinfo {author} {\bibfnamefont {S.}~\bibnamefont
  {Feng}}, \bibinfo {author} {\bibfnamefont {K.}~\bibnamefont {Wang}}, \bibinfo
  {author} {\bibfnamefont {F.}~\bibnamefont {Wang}}, \bibinfo {author}
  {\bibfnamefont {Y.}~\bibnamefont {Zhang}},\ and\ \bibinfo {author}
  {\bibfnamefont {H.}~\bibnamefont {Zhao}},\ }\bibfield  {title} {\bibinfo
  {title} {Less is more: a new machine-learning methodology for spatiotemporal
  systems},\ }\href {https://doi.org/10.1088/1572-9494/ac60f9} {\bibfield
  {journal} {\bibinfo  {journal} {Commun. Theor. Phys.}\ }\textbf {\bibinfo
  {volume} {74}},\ \bibinfo {pages} {055601} (\bibinfo {year}
  {2022})}\BibitemShut {NoStop}%
\bibitem [{\citenamefont {Beck}\ \emph {et~al.}(2024)\citenamefont {Beck},
  \citenamefont {Pöppel}, \citenamefont {Spanring}, \citenamefont {Auer},
  \citenamefont {Prudnikova}, \citenamefont {Kopp}, \citenamefont {Klambauer},
  \citenamefont {Brandstetter},\ and\ \citenamefont {Hochreiter}}]{beck2024}%
  \BibitemOpen
  \bibfield  {author} {\bibinfo {author} {\bibfnamefont {M.}~\bibnamefont
  {Beck}}, \bibinfo {author} {\bibfnamefont {K.}~\bibnamefont {Pöppel}},
  \bibinfo {author} {\bibfnamefont {M.}~\bibnamefont {Spanring}}, \bibinfo
  {author} {\bibfnamefont {A.}~\bibnamefont {Auer}}, \bibinfo {author}
  {\bibfnamefont {O.}~\bibnamefont {Prudnikova}}, \bibinfo {author}
  {\bibfnamefont {M.}~\bibnamefont {Kopp}}, \bibinfo {author} {\bibfnamefont
  {G.}~\bibnamefont {Klambauer}}, \bibinfo {author} {\bibfnamefont
  {J.}~\bibnamefont {Brandstetter}},\ and\ \bibinfo {author} {\bibfnamefont
  {S.}~\bibnamefont {Hochreiter}},\ }\href {https://arxiv.org/abs/2405.04517}
  {\bibinfo {title} {xlstm: Extended long short-term memory}} (\bibinfo {year}
  {2024}),\ \Eprint {https://arxiv.org/abs/2405.04517} {arXiv:2405.04517
  [cs.LG]} \BibitemShut {NoStop}%
\end{thebibliography}%

\end{document}